\documentclass[10pt,twocolumn,letterpaper]{article}

\usepackage{cvpr}              %
\usepackage{multirow}

\usepackage[T1]{fontenc}
\usepackage{mathtools}
\usepackage{algorithm}
\usepackage{csquotes}
\usepackage[noend]{algpseudocode}
\usepackage[table,dvipsnames]{xcolor}
\usepackage{bm}
\usepackage{caption}
\usepackage{xspace}
\usepackage{tikz}
\usepackage{etoolbox}
\usepackage{pgfplots}
\usepackage[numbers]{natbib}
\usepackage{multirow}
\usepackage{booktabs}
\usepackage{dblfloatfix}
\usetikzlibrary{patterns}
\pgfplotsset{compat=1.18}

\newcommand{\dmd}{DMD2-\emph{v}\xspace}
\renewcommand{\gets}{=}
\def\vx{{\bm{x}}}
\def\vy{{\bm{y}}}
\def\vf{{\bm{f}}}

\def\vg{{\bm{g}}}
\def\vm{{\bm{m}}}
\def\vv{{\bm{v}}}
\def\vu{{\bm{u}}}

\makeatletter
\newcommand{\zerodisplayskips}{%
  \setlength{\abovedisplayskip}{2mm}%
  \setlength{\belowdisplayskip}{2mm}%
  \setlength{\abovedisplayshortskip}{1mm}%
  \setlength{\belowdisplayshortskip}{1mm}}
\appto{\normalsize}{\zerodisplayskips}
\appto{\small}{\zerodisplayskips}
\appto{\footnotesize}{\zerodisplayskips}

\renewcommand\paragraph{\@startsection{paragraph}{4}{\z@}%
                                   {1.1ex \@plus0.2ex \@minus0.2ex}%
                                   {-1em}%
                                   {\normalfont\normalsize\bfseries}}

\makeatother

\definecolor{cvprblue}{rgb}{0.21,0.49,0.74}
\usepackage[pagebackref,breaklinks,colorlinks,allcolors=cvprblue]{hyperref}
\usepackage[capitalize]{cleveref}
\Crefname{appendix}{Appendix}{Appendices}

\def\paperID{}
\def\confName{CVPR}
\def\confYear{2026}

\title{Transition Matching Distillation for Fast Video Generation}

\author{
Weili Nie\textonesuperior$^*$,
Julius Berner\textonesuperior$^*$,
Nanye Ma\texttwosuperior,
Chao Liu\textonesuperior,
Saining Xie\texttwosuperior,
Arash Vahdat\textonesuperior \\
\textonesuperior NVIDIA \, \texttwosuperior NYU \, $^*$equal contribution}

\begin{document}
\twocolumn[{%
\renewcommand\twocolumn[1][]{#1}%
\maketitle
\begin{center}
    \centering
    \captionsetup{type=figure}
    \vspace{-0.75em}
    \includegraphics[width=0.95\linewidth]{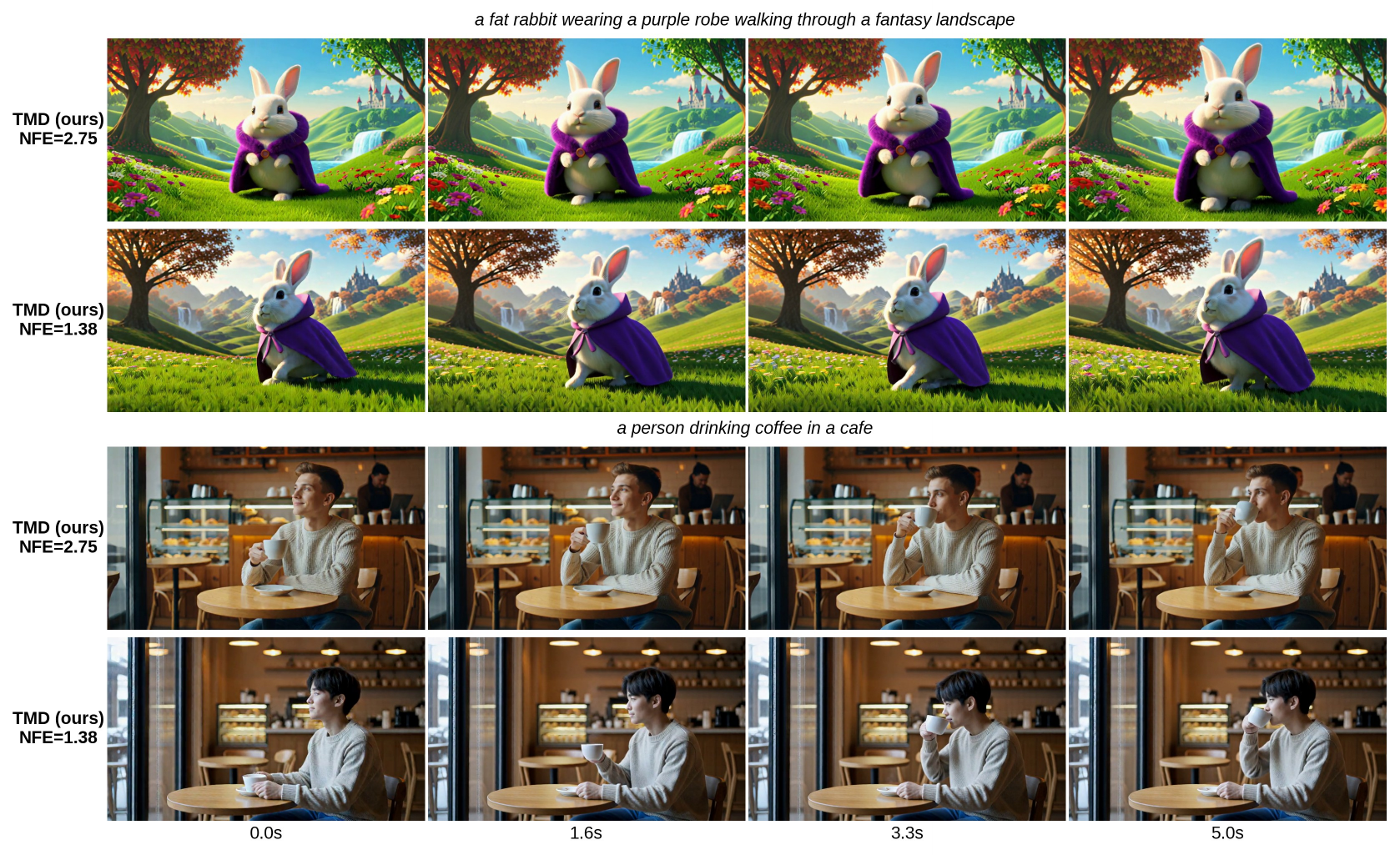}
    \vspace{-0.6em}
    \captionof{figure}{\textbf{Generated examples from TMD.} Four frames of 5s 480p videos generated from two text prompts using our TMD method (distilled from Wan2.1 14B T2V) with two different (effective) number of function evaluations (NFE).}
    \label{fig:example}
\end{center}
}]
\begin{abstract}
\vspace{-20pt}

Large video diffusion and flow models have achieved remarkable success in high-quality video generation, but their use in real-time interactive applications remains limited due to their inefficient multi-step sampling process. In this work, we present Transition Matching Distillation (TMD), a novel framework for distilling video diffusion models into efficient few-step generators. The central idea of TMD is to match the multi-step denoising trajectory of a diffusion model with a few-step probability transition process, where each transition is modeled as a lightweight conditional flow. To enable efficient distillation, we decompose the original diffusion backbone into two components: (1) a main backbone, comprising the majority of early layers, that extracts semantic representations at each outer transition step; and (2) a flow head, consisting of the last few layers, that leverages these representations to perform multiple inner flow updates. Given a pretrained video flow model, we first introduce a flow head to the model, and adapt it into a conditional flow map. We then apply distribution matching distillation to the student model with flow head rollout in each transition step. 
Extensive experiments on distilling Wan2.1 1.3B and 14B text-to-video models demonstrate that TMD provides a flexible and strong trade-off between generation speed and visual quality. In particular, TMD outperforms existing distilled models under comparable inference costs in terms of visual fidelity and prompt adherence.\vspace{-1em}

\end{abstract}
    
\vspace{-0.8em}
\section{Introduction}
\label{sec:intro}
\vspace{-0.1em}

Recent progresses in large-scale diffusion models~\citep{ho2020denoising,song2020score} have significantly advanced the frontier of video generation~\citep{yang2025cogvideox,kong2024hunyuanvideo,videoworldsimulators2024,agarwal2025cosmos,wan2025wan,gao2025seedance}. Open-sourced models (such as HunyuanVideo~\citep{kong2024hunyuanvideo}, Wan~\citep{wan2025wan} and Cosmos~\citep{agarwal2025cosmos}) and commercial text-to-video (T2V) systems (such as Sora,  Veo and Kling) demonstrate remarkable capabilities in synthesizing coherent and photorealistic videos from text prompts. 
Despite their success, sampling inefficiency remains a central bottleneck. Standard diffusion models rely on a multi-step denoising process, often requiring hundreds of iterative steps, to progressively transform noise into realistic outputs~\citep{kong2024hunyuanvideo,wan2025wan}. This iterative nature leads to high inference latency and computational cost, rendering large diffusion models impractical for interactive applications such as real-time video generation, content editing, or world modeling for agent training. Accelerating diffusion sampling without sacrificing visual quality becomes a key open challenge.

A growing body of research has explored diffusion distillation to compress long denoising trajectories into a small number of inference steps.
Existing approaches can be broadly categorized into two families:
(1) \emph{trajectory-based distillation}, which includes knowledge distillation~\citep{luhman2021knowledge,salimans2022progressive} and consistency models~\citep{song2023consistency,geng2024consistency,lu2024simplifying,geng2025mean} that directly regress the teacher’s denoising trajectories; and
(2) \emph{distribution-based distillation}, encompassing adversarial~\citep{sauer2024adversarial,sauer2024fast} and variational score distillation~\citep{yin2024one,zhou2024score,yin2024improved} methods that align the student and teacher distributions.
These techniques can reduce the sampling process to as few as one or two steps in the image domain.
However, extending them to video diffusion models presents unique challenges.
Videos exhibit high spatiotemporal dimensionality and complex inter-frame dependencies, making it difficult to preserve both global motion coherence and fine-grained spatial details during distillation.
Most existing methods treat the diffusion network as a monolithic mapping~\citep{lit20252v,zhang2024sfv,zheng2025rcm}, neglecting the hierarchical structure and semantic progression inherent in large video diffusion backbones.

To address these limitations, we propose \emph{Transition Matching Distillation} (TMD) to distill large video diffusion models into few-step generators (\emph{e.g.}, less than 4 steps; see~\Cref{fig:example} for an example).
Inspired by \emph{Transition Matching}~\citep{shaul2025transition}, TMD approximates the many-step denoising process with a compact few-step probability transition process, where each transition captures the distributional evolution of video samples across widely separated noise levels, enabling the student to take large transition steps that match the teacher model’s distribution.
To model the transition process, we introduce a decoupled architecture for the student model with two components:
(1) a main backbone, comprising the majority of early layers, that extracts high-level semantic representations at each outer transition step; and
(2) a flow head, consisting of the final few layers, that conditions on these representations and refines the fine-grained visual details through multiple inner flow updates.

This hierarchical decomposition allows the student to share representations from the main backbone with the flow head. The flow head performs a few lightweight inner refinement steps within each outer transition step, providing a flexible mechanism to balance sampling efficiency and visual fidelity. First, we pretrain the student via trajectory-based distillation for the flow head using an adaption of \emph{MeanFlow}~\citep{geng2025mean}. Then, we formulate distillation as a distribution matching problem between the teacher denoising process and student transition process using an improved version of \emph{DMD2}~\citep{yin2024improved}.
By unrolling flow head during training, we align the student’s probability transitions with the teacher’s multi-step diffusion distribution, capturing both semantic evolution and fine-grained visual details.

We evaluate TMD in distilling Wan2.1 1.3B and 14B T2V models on VBench~\citep{huang2023vbench} and a user preference study.
Our experiments show TMD consistently outperforms existing distilled methods under comparable inference budgets, achieving better visual fidelity and prompt adherence. Particularly, our distilled 14B model achieves an overall score of 84.24 on VBench in  near-one-step generation (NFE=1.38).
In summary, our main contributions include:
\begin{itemize}[itemsep=0.4em]
    \item \textbf{A novel distillation framework for video diffusion}:
We propose Transition Matching Distillation (TMD), which distills long denoising trajectories into a compact few-step probability transition process.
    \item \textbf{A decoupled diffusion backbone design}:
We decompose the teacher model into a semantic backbone and a recurrent flow head, enabling hierarchical distillation with flexible inner flow refinement.
    \item \textbf{A two-stage training strategy}: (1) transition matching adaption that converts
the flow head into a conditional flow map and (2) distribution matching distillation with flow head rollout in each transition step. 
    \item \textbf{Comprehensive empirical validation}:
We demonstrate the effectiveness of TMD in distilling Wan2.1 1.3B and 14B T2V models, achieving state-of-the-art trade-offs between speed and quality in few-step video generation.
\end{itemize}

\vspace{-6pt}
\section{Background}
\label{sec:background}
\vspace{-2pt}

\begin{figure}[t!]
    \centering
    \includegraphics[width=\linewidth]{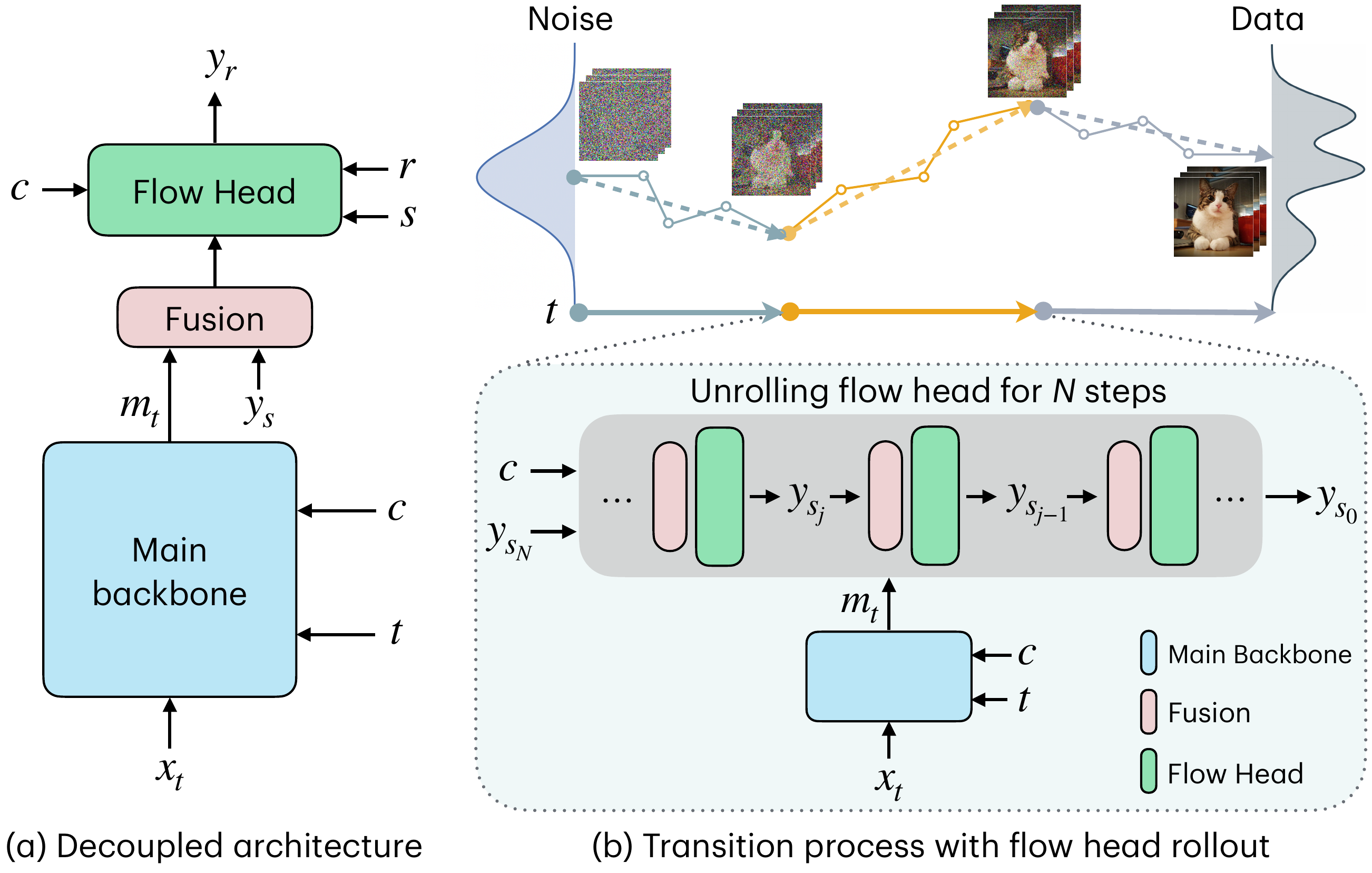}
    \vspace{-14pt}
    \caption{\small \textbf{Overview of our TMD method.} (a) Decoupled architecture for TMD student, where the main backbone takes the noisy sample $\vx_t$, timestep $t$ and text conditioning $c$ as inputs and outputs the main feature $\vm_t$, and with a simple fusion layer, flow head conditions on $\vm_t, c$ and predicts the less noisy target $\vy_r$ from the more noisy $\vy_s$ ($s \geq r$).  (b) Top: Transition process maps noise to data with a few transition steps. Bottom: In each step,  flow head rollout is performed during both distillation and sampling. We omit the timestep inputs $s$ and $r$ to the flow head for simplicity. }
    \label{fig:teaser}
    \vspace{-14pt}
\end{figure}

\paragraph{Transition Matching.}

\emph{Transition matching} (TM)~\citep{shaul2025transition} is a continuous-state generative model, that can be viewed as a generalization of flow matching to discrete time. It uses the rectified flow schedule
\begin{equation}
\label{eq:outer_flow}
    \vx_t = (1-t)\vx + t \vx_1
    , \quad t\in [0,1],
\end{equation}
interpolating between the data $\vx$ and a standard normal $
\vx_1 \sim \mathcal{N}(0,I)$. Flow matching approximates the \emph{instantaneous velocity} 
\begin{equation}
\label{eq:outer_velocity}
    \vv(\vx,t) \coloneqq \mathbb{E}[\Dot{\vx_t}|\vx] = \mathbb{E}[
    \vx_1- \vx|\vx],
\end{equation}
requiring many small time-steps to generate data $\vx$ from noise $\vx_1$~\citep{liu2023flow,albergo2023building,lipman2023flow}. In contrast, TM directly models the probabilistic transition between two states $\vx_{t_{i}}$ to $\vx_{t_{i-1}}$, where $0=t_0 < t_1 < \dots < t_M=1$ is a given time discretization. For every transition the model is learned to predict an auxiliary latent variable $\vy$, such that $\vx_{t_{i-1}}$ is easy to sample (\emph{e.g.}, deterministic) given $\vy$ and $\vx_{t_{i}}$.
For instance, \emph{Difference Transition Matching} (DTM) considers\footnote{While TM uses a probabilistic model to predict $\vy$, flow matching predicts the conditional expectation of $\vy$ (Eq. \eqref{eq:outer_velocity}).} \begin{equation}
\label{eq:dtm}
    \vy \coloneqq \vx_1 - \vx
\end{equation}
such that
\begin{equation}
\label{eq:transition}
    \vx_{t_{i-1}} = \vx_{t_{i}} - (t_{i} - t_{i-1})\vy.
\end{equation}
To predict $\vy$ given $\vx_{t_i}$, transition matching considers an \enquote{inner} rectified flow schedule
\begin{equation}
\label{eq:inner_flow}
    \vy_s = (1-s)\vy + s \vy_1, \quad s\in [0,1],
\end{equation}
where $\vy_1 \sim \mathcal{N}(0,I)$.
In practice, at every step $\vx_{t_i}$, the main backbone predicts features and a lightweight head is learned using flow matching to approximate the inner velocity
\begin{align}
\label{eq:inner_velocity}
    \!\! \vv(\vy,s; \vx_{t_i}) \! \coloneqq \! \mathbb{E}[\Dot{\vy_s}|\vy,\vx_{t_i}] \!
    = \mathbb{E}[\vy_1 \! - \! (\vx_1 \! - \! \vx) |\vy, \vx_{t_i}] \!
\end{align}
While DTM provided improved performance in image generation, it still needs around 30 transition steps in sampling.

\vspace{-3pt}
\paragraph{MeanFlow.} To accelerate the sampling of diffusion models, MeanFlow~\citep{geng2025mean} proposes to learn a \textit{flow map} $\vf(\vy_{s}, s, r)$~\citep{boffi2024flow}, which maps the point $\vy_{s}$ at arbitrary time $s$ on the PF-ODE trajectory to any preceding point $x_r$ for $r < s$. Formally, \begin{align}
\label{eq:flow-map}
\textstyle
    \vf(\vy_{s}, s, r) \coloneqq \vy_{s} - \int_s^r \vv(\vy_\tau, \tau)\,\mathrm{d}\tau
\end{align}
where $\vv$ denotes the instantaneous velocity as in~\eqref{eq:inner_velocity}. MeanFlow parameterizes this mapping using the \textit{average velocity} along the trajectory segment from $\vy_{s}$ to $\vy_r$, expressed as $\vf(\vy_{s}, s, r) = \vy_{s} + (s - r)\vu(\vy_{s}, s, r)$, where $\vu$ represents the average velocity. Combining with~\eqref{eq:flow-map}, we have $(s - r) \vu(\vy_{s}, s, r) = -\int_s^r \vv(\vy_\tau, \tau)\,\mathrm{d}\tau$. This inspires the key insight from MeanFlow, where differentiating both sides w.r.t $s$ obtains the following identity: 
\begin{align}
    \vu(\vy_{s}, s, r) + (s - r) \tfrac{\mathrm{d}}{\mathrm{d}s} \vu(\vy_{s}, s, r) = \vv(\vy_{s}, s),
\end{align}
where $\frac{\mathrm{d}}{\mathrm{d}s} \vu(\vy_{s}, s, r)$ refers to the total derivative. This identity motivates the following practical training objective: 
\begin{align}
\label{eq:mf_objective}
    \mathcal{L}(\theta) \coloneqq
    \mathbb{E}_{s,r,\vy_s}\big[\Vert \vu_\theta(\vy_{s}, s, r) - \hat{\vu} \Vert^2\big]
\end{align}
with 
\begin{equation}\label{eq:jvp}
     \hat{\vu} \coloneqq \text{sg}\big(\vv(\vy_{s}, s) - (s - r)\tfrac{\mathrm{d}}{\mathrm{d}s} \vu_\theta(\vy_{s}, s, r)\big),
\end{equation}
where $\text{sg}(\cdot)$ refers to the stop gradient operator, the instantaneous velocity $\vv(\vy_{s}, s)$ is approximated by conditional velocity $\vy_1 - \vy$, and the total derivative $\frac{\mathrm{d}}{\mathrm{d}s} \vu(\vy_{s}, s, r)$ can be computed using either forward-mode automatic differentiation or a finite-difference approximation~\citep{sun2025unified,wang2025transition}.

\vspace{-3pt}
\paragraph{DMD.} Flow maps like MeanFlow must learn mappings between points along the ODE trajectory, which is difficult to scale to video generation due to high dimensionality and large trajectory curvature (see~\Cref{app:add_exp}).
In contrast, distribution distillation methods such as DMD2~\citep{yin2024improved} offer more flexibility by matching only the output distribution to that of the teacher or data—using a GAN~\citep{goodfellow2020generative} loss for the latter and a variational score distillation (VSD)~\citep{yin2024one} loss, i.e., reverse KL divergence, for the former. This can be represented as
\newcommand{\tdmd}{t}
\begin{equation}
\label{eq:vsd_objective}
\mathcal{L}(\theta) =  \mathbb{E}_{t_i,\vx_{t_i},\tdmd,\hat{\vx}_{\tdmd}}\big[ w(\tdmd) \text{sg}\big(D(\hat{\vx}_{\tdmd},\tdmd) \big)^T \hat{\vx} \big],
\end{equation}
where $\hat{\vx} = \vg_\theta(\vx_{t_i},t_i)$ denotes the student output from the noisy input $\vx_{t_i}$, and $t_i$ is sampled from a given student time discretization, $w$ is a weighting function, $D$ is the difference between the scores of the student and teacher distributions,  $\hat{\vx}_{\tdmd} = \left(1 - \tdmd\right) \hat{\vx} + \tdmd \vx_1$
is a noisy sample from the forward process in~\eqref{eq:outer_flow} that is passed to score functions.

Since we do not have access to the score of the student distribution, a so-called \emph{fake score} is initialized with the teacher parameters and trained on data from the student using flow matching. 
Up to a time-dependent weighting, $D$ equals the difference between the \emph{velocities} of the teacher and fake score. 
In practice, both the fake score and GAN discriminator are updated for a given number of iterations in between updates of the student and the discriminator consists of a lightweight head operating on intermediate features of the fake score or teacher. 

\vspace{-3pt}
\section{Method}
\label{sec:method}

We introduce our method -- \emph{Transition Matching Distillation} (TMD), including two training stages: (1) \emph{transition matching pretraining} to initialize a flow head capable of iteratively refine features extracted from the main backbone; and (2) \emph{distillation with flow head} where we introduce \dmd that advances DMD2 in few-step video generation and apply it with flow head rollout at each transition step. 
For ease of presentation, we drop additional conditioning of the teacher model, such as text conditioning, in our notation. Below, we start with introducing our student architecture before presenting our two-stage training.

\paragraph{Decoupled architecture.} Our method follows the general formulation
of transition matching explained in~\Cref{sec:background}. 
Different from TM, we aim to approximate many small denoising steps of a teacher model with the large transition step of the distilled student. 
To efficiently predict $\vy$ in every transition step ${t_i}$, we decouple the pretrained teacher architecture into a main backbone $\vm_\theta$, that acts as a feature extractor, 
and a lightweight flow head $\vf_\theta$ that iteratively predicts $\vy$ given these features, i.e.,
\begin{equation}
\label{eq:cond_flow_map}
    \vy_{s_{j-1}} \approx \vf_\theta\big(\vy_{s_j}, s_{j}, s_{j-1}; \vm_\theta(\vx_{t_i}, t_i)\big),
\end{equation}
where $0=s_0 < s_1 < \dots < s_N=1$ is a given time discretization for the inner flow; see also~\Cref{fig:teaser}.

While such decoupling has been successfully used for training generative models~\citep{yu2024representation,shaul2025transition,wang2025ddt,lei2025advancing,zheng2025diffusion}, it requires a careful design to minimally disrupt the pretrained model. 
Our design considers two key factors: (1) \emph{flow head target} $\vy$. We find that the DTM formulation $\vy = \vx_1 - \vx$ outperforms other target types, such as the sample prediction $\vy = \vx$ (see~\Cref{app:add_exp}). 
(2) \emph{fusion layer}. We use a time-conditioned gating mechanism to fuse the main feature $\vm_{t_i}$ and the noisy flow head target $\vy_{s_j}$, ensuring the student’s initial forward pass matches that of the teacher. 
See~\Cref{app:implementation_details} for details, where we also provide pseudo-code for our inference in~\Cref{alg:dtm_generation}.

\vspace{-3pt}
\subsection{Stage 1: Transition matching pretraining}

Given the decoupled architecture, we convert the flow head into a flow map for iterative refinement before distillation. 
Similar to TM, we can directly use a flow matching loss in Eq.~\eqref{eq:inner_velocity} to train the flow head to approximate the velocity of the inner flow. However, in theory, this would still require many inner steps to approximate $\vy$. Thus, we leverage MeanFlow to obtain a few-step flow head.

\paragraph{Transition matching MeanFlow.}
On a high-level, our pretraining algorithm, termed \emph{Transition Matching MeanFlow} (TM-MF), uses a MeanFlow objective as in Eq.~\eqref{eq:mf_objective}, conditioned on the main feature $\vm=\vm_\theta(x_{t_i},t_i)$; see~\Cref{alg:dtm_training} for the pseudocode. Specifically, we parametrize the conditional inner flow map using an average velocity as 
\begin{equation}
    \vf_\theta(\vy_{s}, s, r; \vm) \coloneqq \vy_{s} + (s - r)\vu_\theta(\vy_{s}, s, r; \vm).
\end{equation}
Note that we do \emph{not} detach the main backbone features during training as this would limit the flexibility needed in pretraining.
Naively training the flow head to predict the average velocity $\vu_\theta$ performs poorly. Our hypothesis is that the flow head’s output should remain close to the pretrained teacher’s output.
Since the teacher predicts the velocity of the \emph{outer} flow in Eq.~\eqref{eq:outer_velocity}, the flow head should instead predict $\mathbb{E}[\vx_1 - \vx \mid \vx]$ to stay aligned with the teacher.
From the inner velocity definition in Eq.~\eqref{eq:inner_flow}, we obtain
$$\vu_\theta(\vy_{s}, s, s; \vm) \approx \mathbb{E}[\vy_1  - (\vx_1 - \vx) |\vy, \vx_t]$$
Thus, we parametrize the average velocity as
\begin{equation}
\label{eq:tmmf_precond}
    \vu_\theta(\vy_{s},s, r; \vm) \coloneqq \vy_1 - \mathrm{head}_\theta(\vy_{s}, s, r; \vm),
\end{equation}
where $\mathrm{head}_\theta$ is the head of our decoupled architecture (initialized from the teacher as outlined in~\Cref{app:implementation_details}). With this parameterization, the $\mathrm{head}_\theta$ output approximates the teacher velocity prediction in the limit of $r \to s$.

For improved performance and stability, we follow the original MeanFlow to (1) perform flow matching (more precisely, transition matching in our setting) for a part of the batch, (2) use classifier-free guidance (CFG) (adapting the conditional velocity $\vv(\vy_s,s)$), drop the text condition with a certain probability, and (3) use adaptive loss normalization. Since JVP computation in Eq.~\eqref{eq:jvp} requires custom implementations to be compatible with large-scale training for video generation (\emph{e.g.}, flash attention~\citep{dao2022flashattention}, Fully Sharded Data Parallel (FSDP)~\citep{zhao2023pytorch}, or context parallelism~\citep{jacobs2023deepspeed}), we use a finite-difference approximation of the JVP to make our algorithm agnostic of the underlying architectures and training techniques~\citep{sun2025unified,wang2025transition}; see~\Cref{app:implementation_details} for details.

Since we do not have direct access to the inner flow velocity, we use the conditional velocity $\vv(\vy_s,s) = \vy_1 - \vy$ in the objective~\eqref{eq:mf_objective}. We note that for specific $\vy$, one can also derive representations of the inner velocity in terms of the pretrained teacher velocity~\citep{holderrieth2025glass}, which we leave for future work.
Finally, we also observed that TM can provide competitive results as a pretraining strategy; see~\Cref{sec:ablations} for an ablation. In particular, TM pretraining can be understood as a special case of MeanFlow in Eq.~\eqref{eq:mf_objective} for $r=s$ when using the conditional velocity.

\vspace{-4pt}
\subsection{Stage 2: Distillation with flow head}
\label{sec:stage2}
\vspace{-2pt}

After the TM-MF pretraining, we apply distribution distillation to align the student and teacher distributions. We significantly improve the baseline DMD2 for video models (see~\Cref{app:add_exp}) and leverage our optimized implementation for our TMD method. 

\paragraph{DMD2-\emph{v}. } DMD2 was originally proposed for distilling image diffusion models. Thus, its design choices may not be optimal in the video domain. 
We identify three key factors that improve DMD2 for videos (termed DMD2-\emph{v}), which serve as the default setting for our TMD training:
(1) \emph{GAN discriminator architecture}. We find that using Conv3D layers in the GAN discriminator outperforms other architectures, implying the importance of localized, spatio-temporal features for the GAN loss; (2) \emph{Knowledge distillation (KD) warm-up.} We find that KD warm-up improves the performance in one-step distillation, however, in multi-step generation, it tends to introduce coarse-grained artifacts that can be hardly fixed by the DMD2 training (see Figure~\ref{fig:shift_ablation} in Appendix). Therefore, we only apply KD warm-up in one-step distillation for DMD-\emph{v}. 
(3) \emph{Timestep shifting.} When we sample a timestep for the outer transition step or for adding noise to generated samples in the VSD loss, we find that applying a shifting function $t = \frac{\gamma t'}{(\gamma-1)t' + 1}$ with $\gamma \ge 1$ to a uniformly sampled $t'$ improves the performance and prevents mode collapse.

\paragraph{Flow head rollout.} During distillation, we unroll the inner flow and treat the resulting architecture as a sample generator $\vg_\theta(\vx_{t_i},t_i; \vy_1)$ at each transition step $t_i$ (see Figure~\ref{fig:teaser}b).
Considering the flow head target $\vy = \vx_1 - \vx$ defined in Eq.~\eqref{eq:dtm}, the unrolled student output is given by
\begin{equation}
\label{eq:rollout}
   \!\! \vg_\theta(\vx_{t_i},t_i; \vy_1) \!\coloneqq\! \vx_1 -  \textproc{InnerFlow}(\vm_\theta(\vx_{t_i}, t_{i}))
\end{equation}
where $\hat{\vy}_0 \approx \textproc{InnerFlow}(\vm_\theta(\vx_{t_i}, t_{i}))$ denotes the final prediction of the flow head after $N$ inner refinement steps, following Eq.~\eqref{eq:cond_flow_map}.

Applying the VSD loss in Eq.~\eqref{eq:vsd_objective} from DMD2-\emph{v} to the unrolled student output naturally backpropagate gradients through all $N$ inner flow steps. Since the flow head is lightweight, this remains efficient: for instance, if we choose 5 final DiT blocks from a 30-block DiT and unroll 2 steps, it adds less than 17\% extra computation in updating the student network's parameters.
In the terminology of DMD2, this can be viewed as \enquote{backward simulation}~\citep{yin2024improved} of the inner flow, but without detaching the gradient computation. Distillation with flow head rollout effectively avoids the mismatch between training and inference, and thus improves the distillation performance.

\noindent\textbf{Summary:} We provide pseudocode for the student updates for both two stages in~\Cref{alg:dtm_training} of~\Cref{app:implementation_details}.

\begin{table}[t]
\footnotesize
    \centering
    \setlength{\tabcolsep}{3pt}
    \begin{tabular}{l c c c c}
    \toprule
    \multirow{2}{*}{Method} & \multirow{2}{*}{NFE} & Overall & Quality  & Semantic  \\
    & & score  & score  & score  \\
    \midrule    
    \rowcolor{lightgray!20}
    \emph{Diffusion models} & & & & \\
    VideoCrafter2 1.4B~\citep{chen2024videocrafter2} & 50$\times$2 & 80.44 & 82.20 & 73.42 \\
    CogVideoX-2B~\cite{yang2025cogvideox} & 50$\times$2 & 81.55 & 82.48 & 77.81 \\
    Seaweed-7B~\citep{seawead2025seaweed} & 25$\times$2 & 82.15 & 84.36 & 73.31 \\
    Wan2.1 1.3B~\citep{wan2025wan} & 50$\times$2 & {84.26} & {85.30} & {80.09} \\
    \midrule    
    \rowcolor{lightgray!20}
    \emph{Distilled models} & & & & \\
    DOLLAR~\citep{ding2025dollar}$^\ddagger$ & 4 & {82.57} & {83.83} & {77.51} \\
    T2V-Turbo-v2~\citep{lit20252v}$^*$ & 4 & {82.34} & {83.93} & {75.97} \\
    rCM~\citep{zheng2025rcm} & 4 & {84.43} & {85.38} & {80.63} \\
    \dmd & 4 & 84.60 & 86.03 & 79.87 \\
    \hline    APT~\citep{lin2025diffusion}$^\dagger$ & 2 & {81.85} & {84.39} & {71.70} \\    
    rCM~\citep{zheng2025rcm} & 2 & {84.09} & {84.90} & \textbf{80.86} \\
    \dmd & 2 & 84.39 & 85.65 & 79.32 \\
    \dmd & 3 & 84.48 & \underline{85.71} & 79.58 \\
    TMD-N2H5 (Ours) & 2.33 & 
    \textbf{84.68} & \underline{85.71} & \underline{80.55} \\
    TMD-N4H5 (Ours) & 3.00 & \underline{84.67} & \textbf{85.72} & 80.47 \\
    \hline
    APT~\citep{lin2025diffusion}$^\dagger$ & 1 & {82.00} & {84.21} & {73.15} \\ 
    rCM~\citep{zheng2025rcm} & 1 & {82.65} & {83.60} & {78.82} \\
    \dmd & 1 & 83.24 & 84.28 & \textbf{79.10} \\
    TMD-N2H5 (Ours) & 1.17 & 
    \textbf{83.80} & \textbf{85.07} & 78.69 \\
    TMD-N4H5 (Ours) & 1.50 & \underline{83.79} & \underline{85.03} & \underline{78.83} \\

    \bottomrule
    \end{tabular}
    \vspace{-5pt}
    \caption{VBench results of our method and baselines when distilling Wan2.1 1.3B into a few-step generator ($^*$The teacher model is VideoCrafter2 1.4B~\citep{chen2024videocrafter2}; $^\dagger$The teacher model is Seaweed-7B~\citep{seawead2025seaweed}; $^\ddagger$The teacher model is a variant of CogVideoX~\citep{yang2025cogvideox}). \dmd denotes our improved version of DMD2 for video generation. ``N2H5'' (or ``N4H5'') means two (or four) denoising steps and five DiT blocks in the flow head. NFE in our method is represented by the effective forward pass of the whole network as in Eq.~\eqref{eq:nfe}. 
    }
    \label{tab:1_3B_main}
    \vspace{-8pt}
\end{table}

\begin{figure*}[t]
    \centering
    \vspace{-0.2em}
    \begin{minipage}{0.5\linewidth}
        \includegraphics[width=\linewidth]{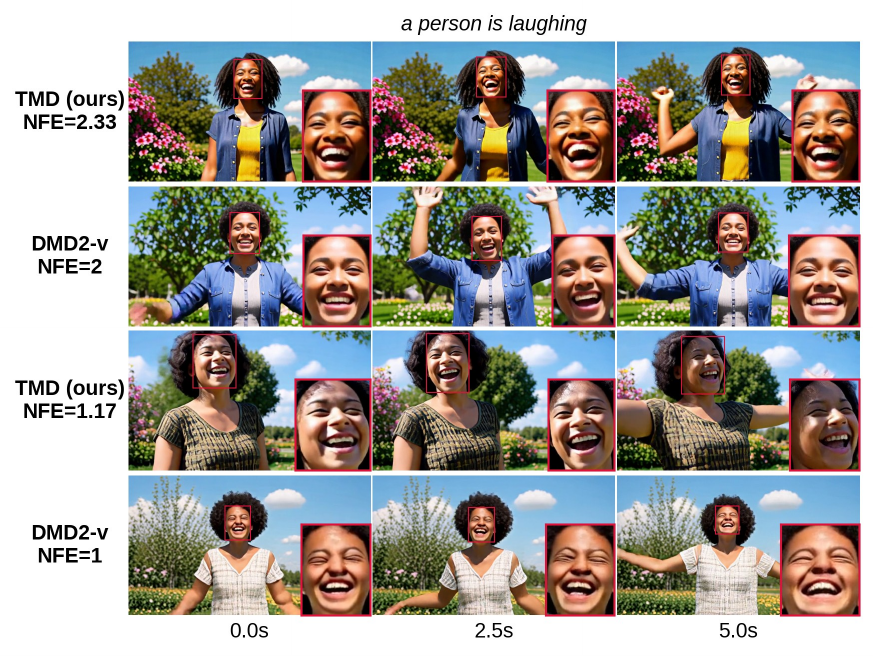}
    \end{minipage}%
    \begin{minipage}{0.5\linewidth}
        \includegraphics[width=\linewidth]{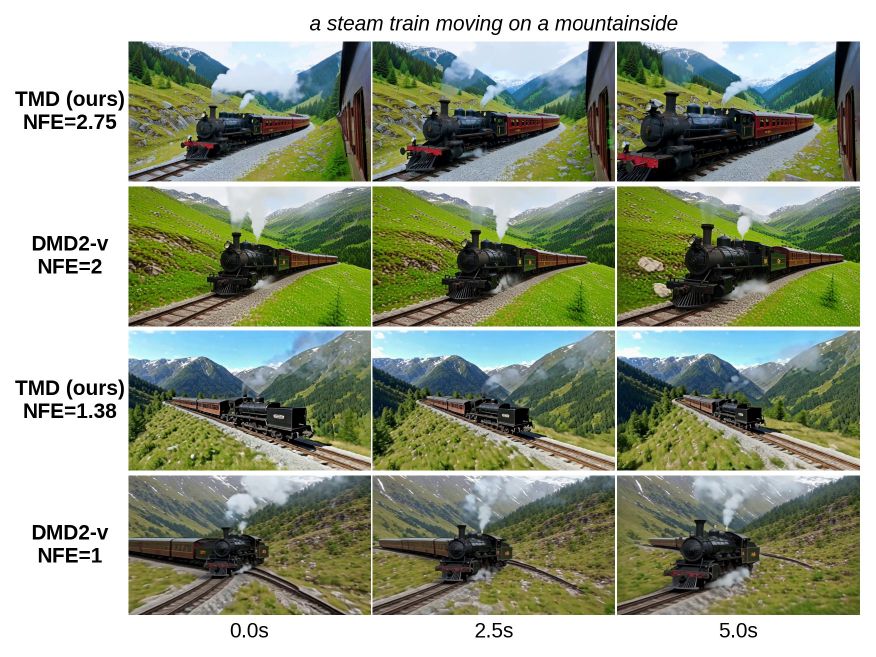}
    \end{minipage}
    \vspace{-0.7em}
    \caption{\textbf{Visual comparison.} We compare three frames (and zoomed-in regions of interest) of the outputs of TMD and \dmd on exemplary prompts for Wan2.1 1.3B (\textbf{left}) and Wan2.1 14B (\textbf{right}). TMD can improve visual quality at comparable cost to our \dmd baseline. 
    Extended prompts can be found in~\Cref{app:implementation_details}.}
    \label{fig:visual_comparison}
    \vspace{-7pt}
\end{figure*}

\vspace{-3pt}
\section{Experiments}
\vspace{-1pt}

\subsection{Experiment setup}
\vspace{-1pt}

\paragraph{Implementation.} We use Wan2.1 1.3B and 14B T2V-480p~\cite{wan2025wan} as teacher video diffusion models and distill them to the same-sized student models with a decoupled architecture.
All experiments are performed on the latent resolution of $[T,H,W] = [21,60,104]$, which gets decoded to 81 frames with pixel resolution $480 \times 832$. We use a dataset of 500k text and video pairs, where text prompts are sampled from the VidProM dataset~\cite{wang2024vidprom} (and extended by Qwen-2.5~\cite{yang2024qwen2}) and videos are generated by the Wan2.1 14B T2V model. We defer more implementation details and hyperparameters to~\Cref{app:implementation_details}.

\paragraph{Evaluation metrics.} To evaluate our method and baselines, we use VBench~\cite{huang2023vbench} (where we report total score, quality score and semantic score) and a user preference study to assess visual quality and prompt adherence. We consider the \emph{effective} number of function evaluations (NFE) to be the total number of DiT blocks used during generation divided by $L$, the number of blocks in the teacher architecture; for the baselines this corresponds to the number of steps $M$, and for our TMD models, this corresponds to
\begin{equation}\label{eq:nfe}
\textstyle
    \text{Effective NFE} \coloneqq M \big( 1 + \frac{(N -1) H} {L} \big),
\end{equation}
where $N$ is the number of inner flow steps and $H$ is the number of blocks in the flow head. We note that $L=30$ for Wan2.1 1.3B and $L=40$ for Wan2.1 14B.

\subsection{Comparison with existing methods}
\label{sec:comparison}

\begin{table}[t]
\footnotesize
    \centering
    \setlength{\tabcolsep}{3pt}
    \begin{tabular}{l c c c c}
    \toprule
    \multirow{2}{*}{Method} & \multirow{2}{*}{NFE} & Overall & Quality  & Semantic  \\
    & & score  & score  & score  \\
    \midrule    
    \rowcolor{lightgray!20}
    \emph{Diffusion models} & & & & \\
    Hunyuan Video 13B~\citep{kong2024hunyuanvideo} & 50$\times$2 & 83.43 & 85.07 & 76.88 \\
    Wan2.1 14B~\citep{wan2025wan} & 50$\times$2 & 86.22 &	86.67 &	84.44 \\
    \midrule    
    \rowcolor{lightgray!20}
    \emph{Distilled models} & & & & \\
    rCM~\citep{zheng2025rcm} & 4 & {84.92} & {85.43} & {82.88} \\
    \dmd & 4 & 84.52 & 85.37 & 81.11 \\
    \hline
    rCM~\citep{zheng2025rcm} & 2 & \textbf{85.05} & \underline{85.57} & \textbf{82.95} \\
    \dmd & 2 & 84.79 & \textbf{85.78} & 80.83 \\
    TMD-N4H5 (Ours) & 2.75 & \underline{84.92} & {85.54} & \underline{82.42} \\
    \hline
    rCM~\citep{zheng2025rcm} & 1 & {83.02} & {83.57} & \underline{80.81} \\
    \dmd & 1 & \underline{83.69} & \underline{84.46} & 80.61 \\
    TMD-N4H5 (Ours) & 1.38 & \textbf{84.24} & \textbf{84.89} & \textbf{81.65} \\
    \bottomrule
    \end{tabular}
    \vspace{-5pt}
    \caption{VBench results of our method and baselines when distilling Wan2.1 14B into a few-step generator. 
    \dmd denotes our improved version of DMD2 for video generation.   
    }
    \label{tab:14B_main}
    \vspace{-8pt}
\end{table}

Our TMD method is built upon the improved version of DMD2 for video generation (termed \dmd). Here, we compare TMD with \dmd and existing baselines in distilling video diffusion models. We provide a visual comparison in \Cref{fig:visual_comparison} and refer to~\Cref{app:add_exp} for further examples. In Table~\ref{tab:1_3B_main}, we show the VBench results of distilling Wan2.1 1.3B (or video models with a similar size) into a few-step generator, where we group the distilled models based on the number of student denoising steps $M$. When $M=2$, TMD-N2H5 with effective $\text{NFE}=2.33$ (i.e., 2 denoising steps and 5 DiT blocks in flow head) obtains an overall score 84.68 that outperforms all other distilled models, including the strongest baseline rCM with $\text{NFE}=4$ (overall score 84.43). When $M=1$, TMD-N2H5 with $\text{NFE}=1.17$ also outperforms all other one-step distillation methods with an overall score 83.80, closing the gap from two-step distillation counterparts. With TMD, we can have more fine-grained control over the quality-efficiency tradeoff by allowing for \textit{fractional} NFEs. 

\begin{figure}[t]
    \centering
    \vspace{-0.5em}
    \includegraphics[width=\linewidth]{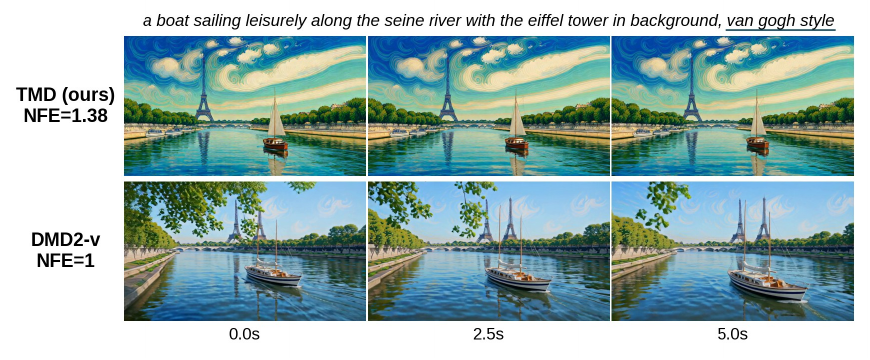}
    \includegraphics[width=\linewidth]{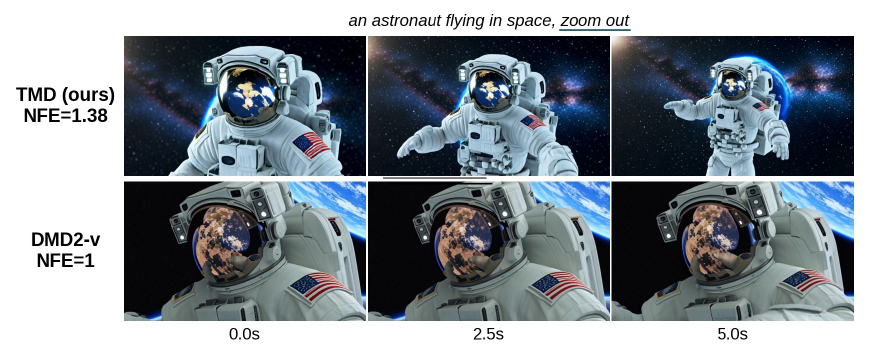}
    \vspace{-1.9em}
    \caption{\textbf{Visual comparison.} We compare three frames of TMD and \dmd on exemplary prompts for Wan2.1 14B. TMD can improve prompt adherence at comparable cost to our \dmd baseline. 
    Extended prompts can be found in~\Cref{app:implementation_details}.}
    \label{fig:prompt_adherance}
    \vspace{-7pt}
\end{figure}

In Table~\ref{tab:14B_main}, we show the VBench results of distilling Wan2.1 14B into a few-step generator. 
When $M=2$, TMD-N4H5 with $\text{NFE}=2.75$ does not outperform 2-step baselines (although it outperforms 4-step DMD-\emph{v}). 
When $M=1$, TMD-N4H5 with $\text{NFE}=1.38$ significantly outperforms all other one-step distillation methods with an overall score of 84.24. It improves over one-step rCM by +1.22 while adding only minimal inference cost. In Figure~\ref{fig:prompt_adherance}, we show TMD improves prompt adherence, supporting the numerical improvement. 
Besides, TMD eliminates the need for the computationally expensive KD warm-up required by one-step \dmd.

\paragraph{User preference study.}

We perform a blinded \emph{two-alternative forced choice} (2AFC) user preference study comparing TMD-N4H5 with DMD2-\emph{v} when distilling Wan2.1 14B under two settings: (1) $M=2$ (two-step generation) and (2) $M=1$ (one-step generation). We randomly sample 60 challenging prompts from VBench~\citep{huang2023vbench} and generate $5$ videos with different seeds for each prompt and model. For each prompt, raters are presented with side-by-side videos from the our model and the baseline (in random order and with random seed) and asked to perform independent pairwise comparisons for two separate criteria: \emph{visual quality} and \emph{prompt alignment}. Detailed instructions are provided in Figure~\ref{fig:user_study}. 
 
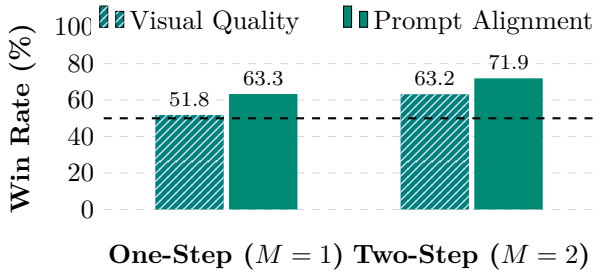
\begin{figure}[t]
    \centering
    \begin{tikzpicture}
        \begin{axis}[
            ybar, 
            bar width=0.9cm, 
            width=\linewidth, 
            height=4cm, 
            enlarge x limits=0.5, 
            ymin=0, ymax=100,
            ymajorgrids=true,
            grid style={dashed, gray!30},
            axis line style={draw=none}, 
            y axis line style={draw=none},
            x axis line style={draw=gray},
            tick style={draw=none},
            ylabel={\textbf{Win Rate (\%)}},
            xtick=data,
            symbolic x coords={One-Step ($M=1$), Two-Step ($M=2$)},
            xticklabel style={font=\bfseries, yshift=-5pt},
            nodes near coords, 
            every node near coord/.append style={font=\footnotesize\bfseries},
            legend style={
                at={(0.5,1.15)}, 
                anchor=north,
                legend columns=-1, 
                draw=none, 
                /tikz/every even column/.append style={column sep=0.5cm}
            }
        ]

        \addplot[
            fill=teal, 
            draw=none, 
            postaction={pattern=north east lines, pattern color=white!30}
        ] coordinates {
            (One-Step ($M=1$), 51.8) 
            (Two-Step ($M=2$), 63.2)
        };

        \addplot[
            fill=PineGreen, 
            draw=none
        ] coordinates {
            (One-Step ($M=1$), 63.3) 
            (Two-Step ($M=2$), 71.9)
        };

        \draw [black, dashed, thick] (rel axis cs:0,0.5) -- (rel axis cs:1,0.5) 
            node [pos=0.02, anchor=south west, font=\scriptsize\bfseries, color=red] {};

        \legend{Visual Quality, Prompt Alignment}
        
        \end{axis}
    \end{tikzpicture}
    \caption{\textbf{User preference study results.} Comparison of TMD-N4H5 (ours) against \dmd under one-step ($M=1$) and two-step ($M=2$) distillation regimes. Values indicate the percentage of times users preferred our method over the baseline \dmd and the dashed line at 50\% represents parity.}
    \label{fig:user_study_results}
\end{figure}

As shown in Figure~\ref{fig:user_study_results}, users consistently preferred TMD over DMD2-\emph{v} in both the one-step and two-step generation settings.
The advantage is even more significant for prompt alignment, echoing the qualitative results in Figure~\ref{fig:prompt_adherance} and underscoring the role of iterative flow-head refinement in largely improving prompt adherence.

\begin{table}[t]
\footnotesize
    \centering
    \begin{tabular}{l c c c}
    \toprule
    \multirow{2}{*}{Discriminator head} & Overall  & Quality  & Semantic  \\
    & score  & score  & score  \\
    \midrule    Conv3D & \textbf{83.24} & \textbf{84.28} & \textbf{79.10} \\
    Conv1D-2D & 82.32 & 83.54 & 77.44 \\
    Attention & 82.36 & 83.32 & 78.51 \\
    \midrule
    w/o GAN & 81.63 & 82.70 & 77.35  \\
    \bottomrule
    \end{tabular}
    \vspace{-5pt}
    \caption{
    Impact of discriminator head design in DMD2-v for one-step distillation of Wan2.1 1.3B. We compare three heads: (1) \emph{Conv3D}, jointly processing spatio-temporal features; (2) \emph{Conv1D–2D}, separating temporal and spatial convolutions (e.g., \cite{zhang2024sfv}); and (3) \emph{Attention}, flattening features into tokens processed by self-attention (with pooling downsampling).
     }
    \label{tab:disc_head}
    \vspace{-5pt}
\end{table}

\begin{table}[t]
\footnotesize
    \centering
    \begin{tabular}{l c c c}
    \toprule
    \multirow{2}{*}{KD warm-up} & Overall  & Quality  & Semantic  \\
    & score  & score  & score  \\
    \midrule 
    One-step w/ KD & \textbf{83.24} & {84.28} & \textbf{79.10} \\
    One-step w/o KD & 83.06 & \textbf{84.72} & 76.42  \\
    \midrule\midrule
    Two-step w/ KD & 83.79 & 84.67 & \textbf{80.28} \\
    Two-step w/o KD & \textbf{84.39} & \textbf{85.65} & 79.32  \\
    \bottomrule
    \end{tabular}
    \vspace{-5pt}
    \caption{Impact of the KD warm-up, where we distill Wan2.1 1.3B into a one-step or two-step student, respectively. For KD warm-up, we use teacher model to generate 10k noise-data pairs. }
    \label{tab:kd}
    \vspace{-3pt}
\end{table}

\begin{table}[t]
\footnotesize
    \centering
    \setlength{\tabcolsep}{1pt}
    \begin{tabular}{l c c c}
    \toprule
    \multirow{2}{*}{Timestep shifting} & Overall  & Quality  & Semantic  \\
    & score  & score  & score  \\
    \midrule 
     $t_{\text{dmd}}$ w/ shift ($\gamma=5$, one-step)  & \textbf{83.24} & {84.28} & \textbf{79.10} \\
    $t_{\text{dmd}}$ w/o shift (one-step)$^*$ & 83.22 & \textbf{84.29} & 78.94  \\
    \midrule\midrule
    $t_{\text{student}}$ w/ shift ($\gamma=10$, two-step) & \textbf{84.39} & \textbf{85.65} & \textbf{79.32} \\
    $t_{\text{student}}$ w/o shift (two-step) & 83.44 & 84.96 & 77.33  \\
    \bottomrule
    \end{tabular}
    \vspace{-5pt}
    \caption{Impact of the timestep shifting for $t_\text{dmd}$ that controls the noise level in the DMD loss and $t_\text{student}$ that controls the denoising steps in the multi-step student, where we distill Wan2.1 1.3B into a one-step or two-step student, respectively. $^*$Note that the setting ``$t_{\text{dmd}}$ w/o shift'' leads to the severe mode collapse that VBench scores cannot capture (see~Fig. \ref{fig:shift_ablation} in Appendix).}
    \label{tab:t_shift}
    \vspace{-3pt}
\end{table}

\vspace{-1pt}
\subsection{Ablation studies}
\label{sec:ablations}

\paragraph{Three design choices in DMD2-\emph{v}. }
Tables~\ref{tab:disc_head}-\ref{tab:t_shift} show the impact of the discriminator head, KD warm-up and timestep shifting in DMD2-\emph{v}, respectively. 
In Table~\ref{tab:disc_head}, we observe that adding the GAN loss improves the distillation performance, and \emph{Conv3D} outperforms the other two discriminator head architectures. 
In Table~\ref{tab:kd}, the overall score on VBench increases with KD warm-up in one-step DMD2 but decreases with KD warm-up in two-step DMD2. It implies that we better only apply KD warm-up in one-step generation. In Table~\ref{tab:t_shift}, we observe that applying timestep shifting to $t_\text{dmd}$ that controls the
noise level in the DMD loss and $t_\text{student}$ that controls the denoising
steps in the multi-step student improves the distillation performance, respectively. 

\begin{figure}[t]
    \centering
    \includegraphics[width=0.85\linewidth]{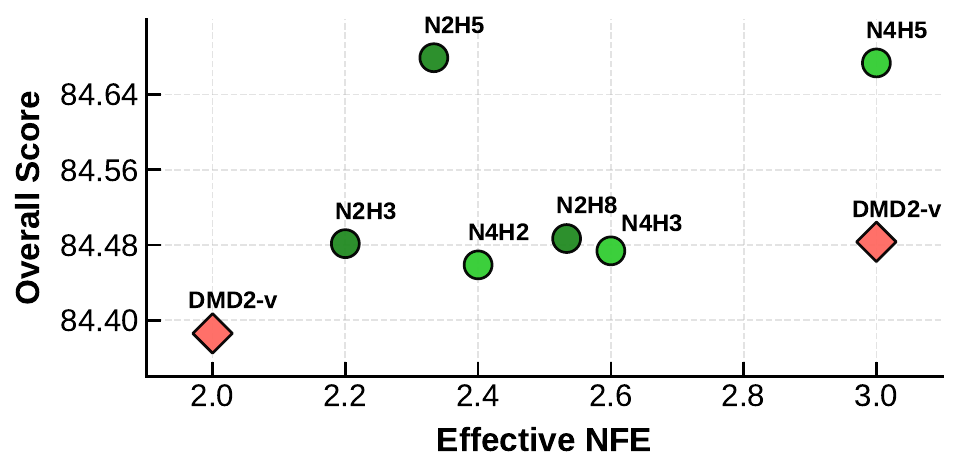}
    \vspace{-0.8em}
    \caption{\textbf{Performance-efficiency tradeoff of TMD.} We compare the overall VBench score and effective NFE of TMD, when distilling Wan2.1 1.3B with $M=2$ for different number of inner steps $N$ and flow head layers $H$ against $2$- and $3$-step \dmd. TMD can provide consistent performance gains for increasing NFE.}
    \vspace{-0.5em}
    \label{fig:tradeoff}
\end{figure}

\paragraph{Quality-efficiency tradeoff. } 
The number of inner steps $N$ and flow head layers $H$ control the computational cost of inner flows.
We vary $N$ and $H$ to more comprehensively analyze TMD's performance-efficiency tradeoff. In Figure~\ref{fig:tradeoff}, we observe that the overall VBench score generally improves as the effective NFE increases. This justifies the fine-grained flexibility that our method offers in balancing generation speed and visual quality. 

\paragraph{MeanFlow \emph{vs.\@} flow matching.}
In transition matching pretraining, we replace the MeanFlow objective with the vanilla flow matching objective (TM) to highlight the impact of MeanFlow (TM-MF). \Cref{tab:pretraining} shows that TM-MF consistently achieves better distillation performance than TM, suggesting that TM-MF offers a superior initialization for the second-stage distillation training.

\begin{table}[t]
\footnotesize
    \centering
    \begin{tabular}{l l c c c}
    \toprule
    \multirow{2}{*}{Setting} & \multirow{2}{*}{Pretraining} & Overall  & Quality  & Semantic  \\
    & & score  & score  & score  \\
    \midrule    
    \multirow{2}{*}{N2H5}  & TM-MF & \textbf{84.68} & \textbf{85.71} & \textbf{80.55} \\
      & TM & 84.61 & 85.64 & 80.46 \\
    \midrule
    \multirow{2}{*}{N4H5}  & TM-MF & \textbf{84.67} & \textbf{85.72} & \textbf{80.47} \\
    & TM & 84.29 & 85.45 & 79.68 \\
    \bottomrule
    \end{tabular}
    \vspace{-5pt}
    \caption{Impact of the first stage pretraining method on the final performance when we distill Wan2.1 1.3B with $M=2$.}
    \label{tab:pretraining}
    \vspace{-10pt}
\end{table}

\begin{figure}[t]
    \centering
    \includegraphics[width=0.85\linewidth]{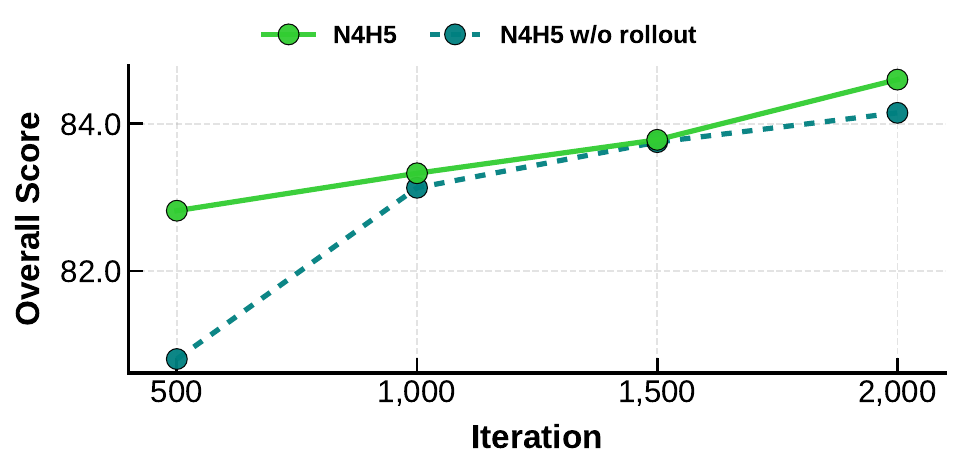
    }
    \vspace{-0.8em}
    \caption{\textbf{Convergence and rollout ablation.} We compare the overall VBench score over iterations for the second-stage TMD training with and without flow head rollout. While TMD generally converges within a only few thousand iterations, we observe faster convergence and improved performance when using rollouts.}
    \vspace{-0.9em}
    \label{fig:rollouts}
\end{figure}

\paragraph{Flow head rollout in distillation. }
It is crucial to close the gap between training and inference, by allowing gradients from the distillation objective to backpropagate through the unrolled inner flow trajectory. \Cref{fig:rollouts} shows that applying flow head rollout in distillation largely leads to faster training convergence and improved performance.

\vspace{-6pt}
\section{Related works}
\vspace{-3pt}

\paragraph{Accelerating diffusion models.} 
Recent efforts to accelerate diffusion models for few-step generation generally follow two directions:
Trajectory matching methods directly learn mappings along the underlying ODE trajectory~\cite{salimans2022progressive,song2023consistency,geng2025mean,sun2025unified,zheng2023fast,boffi2024flow,frans2024one,lu2024simplifying,geng2024consistency,lee2024truncatedconsistencymodels,sabour2025align,piflow}, while distribution matching methods train a student model to match the teacher’s distribution with fewer steps~\cite{sauer2024fast,yin2024improved,xu2025one,yin2024one,sauer2024adversarial,zhou2024score,luo2024one}.
In trajectory matching, Progressive Distillation~\cite{salimans2022progressive} and Shortcut Models~\cite{frans2024one} gradually reduce sampling steps during training, while Consistency Models~\cite{song2023consistency,lu2024simplifying,geng2024consistency} directly map intermediate states to clean data.  Subsequent works~\cite{geng2025mean,sabour2025align,frans2024one,boffi2024flow,kim2023consistency} extend this idea to learn mappings (or \emph{flow maps}) between any pair of timesteps. They often face training instability when scaled to larger models or higher resolutions~\cite{lu2024simplifying,chen2025sanasprint,zheng2025rcm}.
Distribution matching approaches, in contrast, align the student’s few-step sampling distribution with the teacher’s via distributional divergence~\cite{yin2024one,yin2024improved,xu2025one}. This provides stable supervision at the distribution level and is less sensitive to the curvature of ODE trajectories, with adversarial objectives further improving sample fidelity.

As video generation grows in prominence along with its much higher sampling cost, recent works have extended these acceleration techniques to the video domain~\cite{zheng2025rcm,zhang2024sfv,lin2025diffusion,ding2025dollar}. Our method advances this line by introducing a decoupled architecture that more effectively leverages the teacher’s learned semantic representations.

\paragraph{Efficient video generation.} 
Beyond reducing sampling steps, other research aims to enhance the overall efficiency of video diffusion models.
Autoregressive–diffusion hybrids achieve this by training causal models that generate frames sequentially instead of predicting the entire sequence at once~\cite{yin2025slow,huang2025self,cui2025self,lin2025autoregressive,teng2025magi,chen2024diffusion,zhang2025packing}. This design significantly reduces attention complexity per forward pass through shorter temporal contexts.
Meanwhile, system-level optimizations such as feature caching and efficient attention further reduce sampling cost by reusing intermediate activations~\cite{liu2024faster,zhao2024real,ma2024learning} or by sparsifying or linearizing the attention operation~\cite{yang2025sparse,xi2025sparse,chen2025sanavideo}. These techniques are complementary to ours and can be combined with TMD for even greater video generation efficiency.

\paragraph{Decoupled backbone.}
Recent studies have demonstrated the advantages of decoupling diffusion backbones into a feature extractor (encoder) and a head (decoder)~\cite{yu2024representation,shaul2025transition,wang2025ddt,lei2025advancing,zheng2025diffusion}.
For example, aligning intermediate features with those from pretrained visual encoders improves training convergence and condition adherence~\cite{yu2024representation,bhowmik2025moalign,zhang2025videorepa}.
Decoupled Diffusion Transformer (DDT)~\cite{wang2025ddt} further accelerates sampling by reusing extracted features across denoising steps, though such encoder sharing is not explicitly enforced during training.
Transition Matching (TM)~\cite{shaul2025transition} explicitly models an inner flow loop for the head via flow matching~\citep{lipman2022flow}.
Our work extends TM in two key aspects: (1) scaling it to large-scale video generation, and (2) distilling pretrained video diffusion models into few-step generators instead of training a multi-step TM model from scratch.

\vspace{-5pt}
\section{Conclusions}
\vspace{-3pt}

We introduced Transition Matching Distillation (TMD), a novel framework of addressing the significant inference latency of large-scale video diffusion models. The core of our method lies in a decoupled student architecture, which separates a main backbone for semantic feature extraction from a lightweight, recurrent flow head for iterative refinement. This design is coupled with a two-stage training strategy involving transition matching pretraining and distribution-based distillation.
Our experiments on distilling Wan2.1 models showed that TMD offers fine-grained flexibility in balancing the trade-off between generation speed and video quality. TMD can outperform existing distillation techniques, achieving superior visual fidelity and prompt adherence at comparable or even reduced computational budgets. 
Future directions include (1) unifying the two training stages into a single-stage pipeline and (2) integrating TMD with system-level optimizations, such as efficient attention or feature caching, to further accelerate video generation.
\par

{
    \small
    \bibliographystyle{ieeenat_fullname}
    \bibliography{main}
}

\crefalias{section}{appendix}
\clearpage

\appendix

\begin{table*}[t] 
\centering 
\renewcommand{\arraystretch}{1.2} 
\setlength{\aboverulesep}{0pt}
\setlength{\belowrulesep}{0pt}
\vspace{-5pt}
\resizebox{0.85\textwidth}{!}{
\begin{tabular}{lcc} 
\toprule 
\rowcolor[HTML]{EFEFEF} 
\textbf{Hyperparameter} & \textbf{Wan2.1 1.3B (T2V)} & \textbf{Wan2.1 14B (T2V)} \\ 
\midrule 
\multicolumn{3}{c}{\textbf{General Settings}} \\ 
\textbf{Resolution} &  \multicolumn{2}{c}{$480 \times 832$ (480P)}  \\ 
\textbf{Frame count} & \multicolumn{2}{c}{$81$ Frames ($5$s)} \\
\textbf{Latent dimension $(T\times H \times W)$} &  \multicolumn{2}{c}{$21 \times 60 \times 104$} \\
\textbf{Dataset size} & \multicolumn{2}{c}{500k (479k after filtering)} \\
\textbf{Dataset prompts} &  \multicolumn{2}{c}{VidProm extended by \textsc{Qwen2.5-7B}} \\ 
\textbf{Dataset videos} &  \multicolumn{2}{c}{Generated by Wan2.1 14B (T2V)} \\ 
\textbf{Optimizer (weight decay, betas, epsilon)} &  \multicolumn{2}{c}{AdamW ($0.01$, $(0.9,0.99)$, $10^{-8}$)} \\ 
\textbf{Global batch-size} & \multicolumn{2}{c}{64} \\
\textbf{Timestep shift $\gamma$ for $t_{\text{student}}$} (see~\Cref{sec:stage2}) & \multicolumn{2}{c}{10} \\
\textbf{Multi-step sampling} (see~\Cref{app:dmd2}) & \multicolumn{2}{c}{deterministic} \\
\textbf{Precision} & \multicolumn{2}{c}{BF16 (time $t$ in FP64)} \\
\textbf{GPU type} &  NVIDIA A100 & NVIDIA H100 \\ 
\textbf{Parallelism Strategy} &  DDP & FSDP \\ 
\midrule
\rowcolor[HTML]{EFEFEF} 
\multicolumn{3}{c}{\textbf{Baselines}} \\ 
\midrule
\multicolumn{3}{c}{\textbf{KD}} \\ 
\midrule
\textbf{Number of trajectories} & \multicolumn{2}{c}{10k} \\
\textbf{CFG scale} & \multicolumn{2}{c}{5} \\
\textbf{Skip-layer} & \multicolumn{2}{c}{10} \\
\textbf{Learning rate} & \multicolumn{2}{c}{7e-5} \\
\textbf{Max. iterations} & 10k & 5k \\
\midrule
\multicolumn{3}{c}{\textbf{\dmd}} \\ 
\midrule
\textbf{Fake score architecture} & \multicolumn{2}{c}{same as teacher} \\
\textbf{Discriminator architecture} (see~\Cref{sec:ablations}) & \multicolumn{2}{c}{Conv3D} \\
\textbf{Discriminator parameters} & {68M} & 172M \\
\textbf{CFG scale} & \multicolumn{2}{c}{5} \\
\textbf{Layer for discriminator features} (see~\Cref{app:dmd2}) & $(15,22,29)$ & $(19,29,39)$ \\
\textbf{Discriminator loss weight $\lambda$} (see~\Cref{alg:dtm_training}) & \multicolumn{2}{c}{0.03} \\
\textbf{Learning rates (student, discriminator, fake score)} & \multicolumn{2}{c}{$(10^{-5},10^{-5},10^{-5})$} \\
\textbf{Timestep shift $\gamma$ for $t_{\text{dmd}}$} (see~\Cref{sec:stage2}) & \multicolumn{2}{c}{5} \\
\textbf{Min./max. for $t_{\text{dmd}}$ and $s_{\text{mf}}$} & \multicolumn{2}{c}{$(0.001,0.999)$} \\
\textbf{Student update frequency} & \multicolumn{2}{c}{every $5$-th iteration}  \\
\textbf{Max. iterations} & $4$k $(M \ge 2)$, $12$k $(M=1)$ & $1$k $(M \ge 2)$, $5$k $(M=1)$  \\
\midrule
\rowcolor[HTML]{EFEFEF} 
\multicolumn{3}{c}{\textbf{TMD}} \\ 
\midrule
\multicolumn{3}{c}{\textbf{Stage 1: TM-MF}} \\ 
\midrule 
\textbf{CFG scale} & \multicolumn{2}{c}{3} \\
\textbf{Flow head fuse type} (see~\Cref{app:fuse}) & \multicolumn{2}{c}{gated} \\
\textbf{Timestep shift $\gamma$ for $s_{\text{student}}$} (see~\Cref{app:tmmf}) & \multicolumn{2}{c}{10} \\
\textbf{Timestep shift $\gamma$ for $s_{\text{mf}}$} (see~\Cref{app:tmmf}) & \multicolumn{2}{c}{3} \\
\textbf{JVP finite-difference $\delta$} (see~\Cref{app:tmmf}) & \multicolumn{2}{c}{$5\cdot 10^{-3}$} \\
\textbf{Condition dropout probability} (see~\Cref{app:tmmf}) & \multicolumn{2}{c}{$0.1$} \\
\textbf{Loss normalization constant $c$} (see~\Cref{app:tmmf}) & $d$ & $10^{-5}d$ \\
\textbf{Learning rate} & $3\cdot 10^{-5}$ & $1\cdot 10^{-5}$ \\
\textbf{Max. iterations} & 3k & 3k \\
\midrule 
\multicolumn{3}{c}{\textbf{Stage 2: \dmd with flow head}} \\ 
\midrule 
\multicolumn{3}{c}{All hyperparameters as in \dmd above} \\
\bottomrule 
\end{tabular}
}
\vspace{-5pt}
\caption{Default hyperparameters if not specified otherwise in the experiments.
} 
\label{tab:hp} 
\vspace{-5pt}
\end{table*}

\section{Implementation details}
\label{app:implementation_details}

\subsection{Flow head conditioning}
\label{app:fuse}

\paragraph{Time conditioning.}

For the flow head, we re-use the default time embedding to embed $s$ and instantiate an additional zero-initialized time embedding that embeds the difference $s-r$ between two time steps of the flow map (see~\Cref{sec:background}). The two embeddings are summed and used as the conditioning input to the adaptive normalization layers in the DiT blocks of the flow head.

\paragraph{Inner flow fusion.}

\begin{figure}[t]
    \centering
    \includegraphics[width=\linewidth]{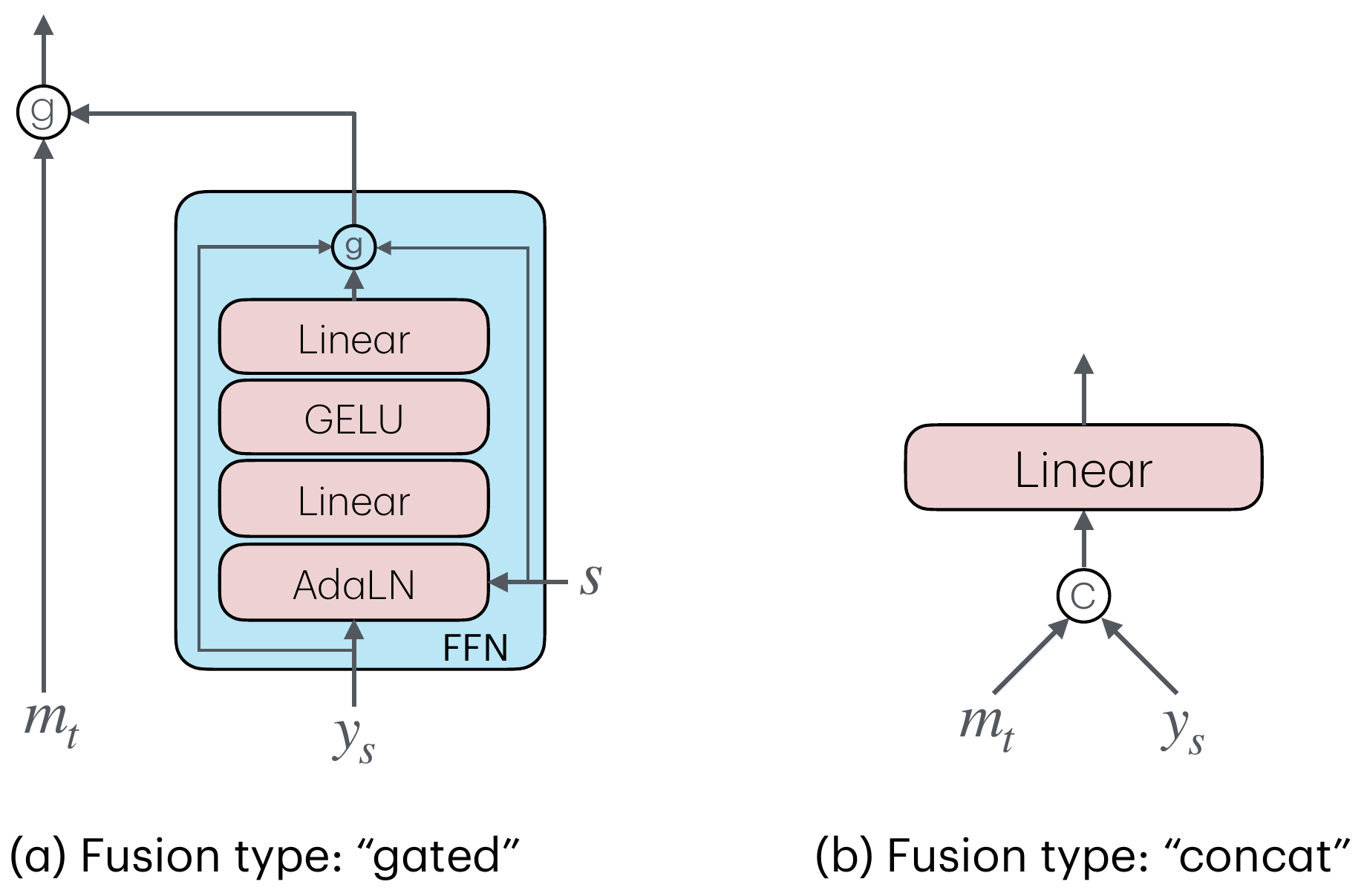}
    \caption{Visualization of our mechanisms to fuse the main backbone's features $\vm(\vx,t)$ with the flow head input $\vy_s$: (a) ``gated'' fusion, where the flow head input $\vy_s$ is passed to an FFN block (conditioned on the flow head timestep $s$) and then added to the main feature $\vm_t$ via a gated operator;  and (b) ``concat'' fusion, where the the flow head input $\vy_s$  and the main feature $\vm_t$ are concatenated in the channel dimension and then passed to the linear projection layer.}
    \label{fig:gating}
\end{figure}

To fuse the main backbone's features $\vm(\vx,t)$ with the flow head input $\vy_s$, we first process $\vy_s$ with the patch embedding layer of the main backbone. Then we process the patches using a randomly initialized AdaLN-style block (analogous to the ones used in the teacher model, \emph{i.e.}, consisting of adaptive layer normalization, an MLP, and a gated residual connection) conditioned on the embedding of $s$ and re-using the global scale and shift values. Finally, we fuse this processed flow head input with the main backbone's features using a gated interpolation, controlled by a global gate derived from a sigmoid-activated learnable parameter. This makes the flow head's contribution both time-aware and dynamically gated at the feature level before the final global fusion. We visualize the fusion mechanism in~\Cref{fig:gating} and provide an ablation w.r.t.\@ other conditioning types in~\Cref{app:fuse_ablation}.

\begin{algorithm}[t]
\footnotesize
\caption{TMD inference}
\label{alg:dtm_generation}
\begin{algorithmic} 
\State $\vx \sim \mathcal{N}(0, I)$
\For{$i = M$ \textbf{to} $1$} 
\State $\vm \gets \vm_\theta(\vx, t_{i})$ \Comment{Main backbone}
\State $\vx \gets \vx - (t_{i} - t_{i-1})\textproc{InnerFlow}(\vm)$ \Comment{See~Eq.~\eqref{eq:transition}}

\EndFor

\State \Return $\vx$ \Comment{Generated data}
\vspace{1em}
\Procedure{InnerFlow}{$\vm$}
    \State $\vy \sim \mathcal{N}(0, I)$ 
    \For{$j = N$ \textbf{to} $1$} 
        \State $\vy \gets \vf_\theta(\vy, s_{j}, s_{j-1}; \vm)$ \Comment{Flow head}
    \EndFor
    \State \Return $\vy$
\EndProcedure
\end{algorithmic}
\end{algorithm}

\subsection{TM-MF}
\label{app:tmmf}

\paragraph{Pretraining recipe.} To better align TM-MF pretraining with the multi-step inference strategy used in our distillation setup, we modify the sampling scheme of $(s, r)$ compared to the continuous formulation in~\cite{geng2025mean}. First, analogous to $t_{\mathrm{student}}$ in the outer flow of \dmd, we also use a shifting function with $\gamma=10$ for $s_{\mathrm{student}}$, i.e., the values defining the time grid\footnote{Note that, in practice, we set $t_M = s_N = 0.999$ instead of $1$ to align with the pretraining of Wan.} $0=s_0 < s_1 < \dots < s_N=1$. Secondly, analogous to sampling the timestep $t_{\mathrm{dmd}}$ in \dmd, we also sample $s_{\mathrm{mf}}$ during TM-MF training uniformly and shift it (using $\gamma=3$). Then, we pick $r=s_{k}$ where $k\coloneqq \max \{j: s_{j} \le s_{\mathrm{mf}} \}$. Empirically, this discrete sampling scheme is not only tailored to our Stage 2 distillation but also substantially stabilizes TM-MF pretraining. 

Our remaining design choices largely follow~\citet{geng2025mean}. We set $r = s$ for $75\%$ of training batches to stabilize optimization under the flow-matching loss, apply condition dropout (using the negative prompt in~\Cref{tab:ext_prompts}) during training, and construct the velocity field using classifier-free guidance. We observed that mixing conditional and unconditional network-predicted velocities (Eq.\@ (18) in~\citep{geng2025mean}) offers no clear benefit in our setting, so we adopt the standard classifier-free guidance formulation throughout. Moreover, we also use an adaptive loss normalization, such that our final loss derived from~\eqref{eq:mf_objective} is given by
\begin{equation*}
    \mathbb{E}_{s,r,\vy_s}\left[\frac{\Vert \vu_\theta(\vy_{s}, s, r) - \hat{\vu} \Vert^2}{\operatorname{sg}\big(\Vert \vu_\theta(\vy_{s}, s, r) - \hat{\vu} \Vert^2\big) + c}\right],
\end{equation*}
where we choose $c=d$ for Wan2.1 1.3B and $c=\frac{d}{10^5}$ for Wan2.1 14B (where $d$ is the dimension of $\vy_s$).

\begin{algorithm}[t]
\footnotesize
\caption{TMD student update step (simplified)}
\label{alg:dtm_training}
\begin{algorithmic} 

\State Given $\vx \sim p_{\mathrm{data}},\ \vx_1 \sim \mathcal{N}(0,I),\ t_{i} \sim \mathrm{Unif}(\{t_1,\dots,t_M\})$
\State $\vx_{t_i} \gets (1-t_i)\vx + t_i \vx_1$ \Comment{See Eq.~\eqref{eq:outer_flow}}
\State $\vm \gets \vm_\theta(\vx_{t_i}, t_{i})$ \Comment{Main backbone}
\If{stage\_one}
\State $\vy \gets \vx_1 - \vx$ \Comment{See Eq.~\eqref{eq:dtm}}
\State $\vy_1\sim \mathcal{N}(0,I),\ (s,r) \sim p_{s,r}$ 
\State $\vy_s \gets (1-s)\vy + s \vy_1$ \Comment{See Eq.~\eqref{eq:inner_flow}}
\State $\vu \gets \vu_\theta(\vy_s, s, r; \vm)$ \Comment{Avg.\@ velocity}
\State $\vv \gets \vy_1-\vy$ \Comment{Conditional velocity}
\State $\mathcal{L} \gets \text{MeanFlow}(\vu,\vv,s,r)$ \Comment{See Eq.~\eqref{eq:mf_objective}} 
\Else
\State $\hat{\vx} \gets \vx_1 - \textproc{InnerFlow}(\vm)$ \Comment{See Eq.~\eqref{eq:rollout} \& \Cref{alg:dtm_generation}}
\State $\mathcal{L} \gets \mathrm{VSD}(\hat{\vx}) + \lambda \cdot \mathrm{Discriminator}(\hat{\vx})$ \Comment{See Eq.~\eqref{eq:vsd_objective}}
\EndIf
\State $\theta \gets \mathrm{step}(\theta,\nabla_\theta \mathcal{L})$ \Comment{Gradient step}
\end{algorithmic}
\end{algorithm}

\paragraph{Finite difference approximation.} Computing the JVP term in the MeanFlow objective, namely $\tfrac{\mathrm{d}}{\mathrm{d}s} \vu_\theta(\vy_{s}, s, r)$, using forward-mode automatic differentiation in PyTorch is currently incompatible with system optimizations, such as flash attention and FSDP; however, these optimizations are crucial to avoid out-of-memory issues due to the long video sequence length. To address this and make our method agnostic of different training techniques, we approximate the JVP using a \emph{central difference scheme}. For a step size $\delta$, we have 
\begin{align*}
    \tfrac{\mathrm{d}}{\mathrm{d}s} &\vu_\theta(\vy_{s}, s, r) \approx\frac{\vu_\theta(\vy_{s + \delta}, s + \delta, r) - \vu_\theta(\vy_{s - \delta}, s - \delta, r)}{2\delta}  
\end{align*}
with $\vy_{s \pm \delta} = \vy_s \pm \delta\vv(\vy_s, s)$, where we use the conditional velocity for $\vv$. At the boundaries, we fall back to using one-sided finite-difference approximations. We empirically found that setting $\delta = 0.005$ yields a satisfactory estimation to the JVP term and fix it throughout our experiments.

\begin{figure}[t!]
    \centering
    \includegraphics[width=\linewidth]{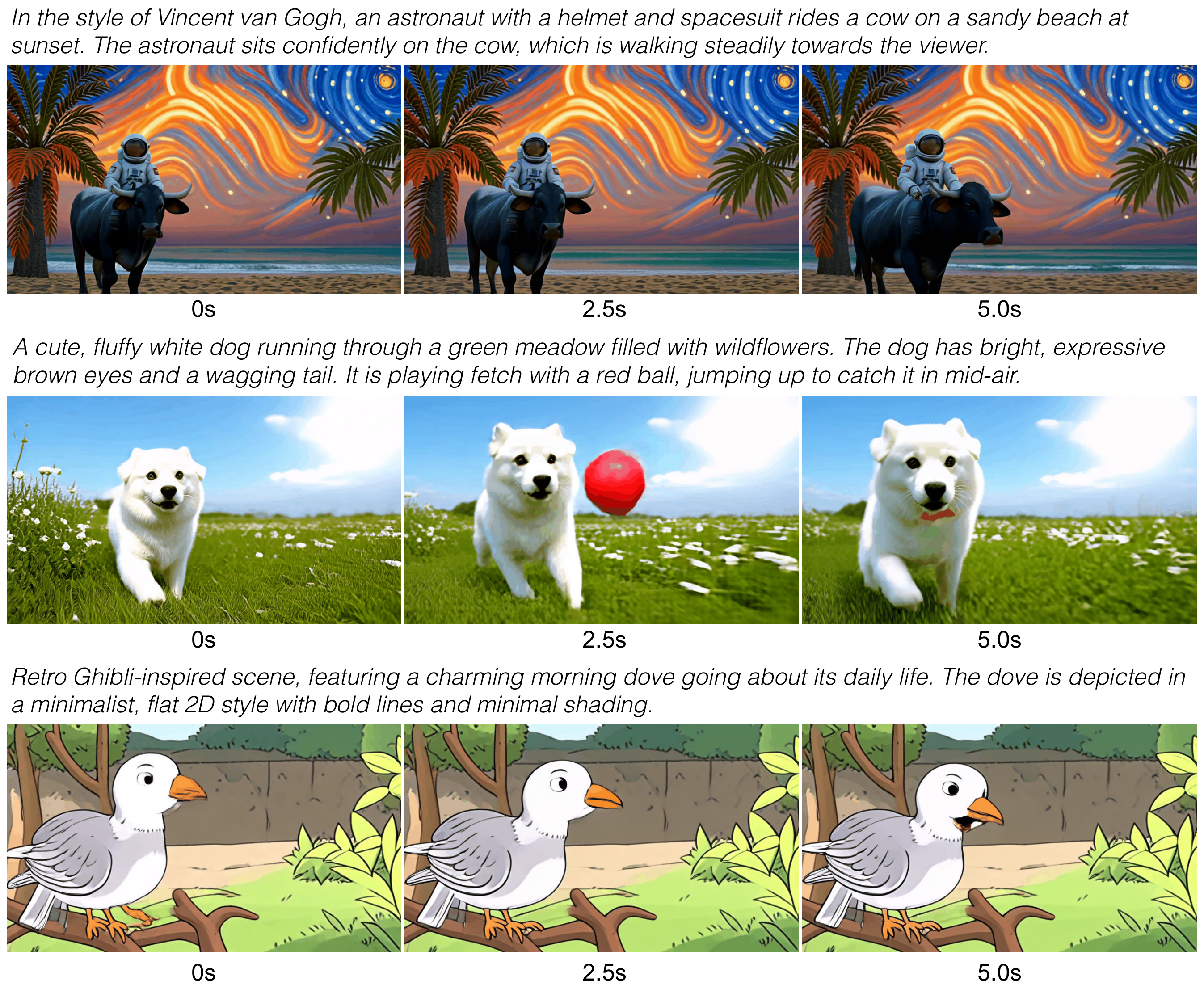}
    \vspace{-12pt}
    \caption{\small 
    \textbf{Mode collapse without time-shifting.} We  show videos generated by the one-step student distilled from DMD2 in the setting ``$t_{\text{dmd}}$ w/o shift'' (see Table~\ref{tab:t_shift}), where the other hyperparameters use the default values. 
    We can see that all generated videos have the main characters consistently appear on the left side of the pixel space, which is a sign of the severe mode collapse happening during distillation.}
    \label{fig:kd_ablation}
    \vspace{-5pt}
\end{figure}

\begin{figure*}[t!]
    \centering
    \includegraphics[width=\linewidth]{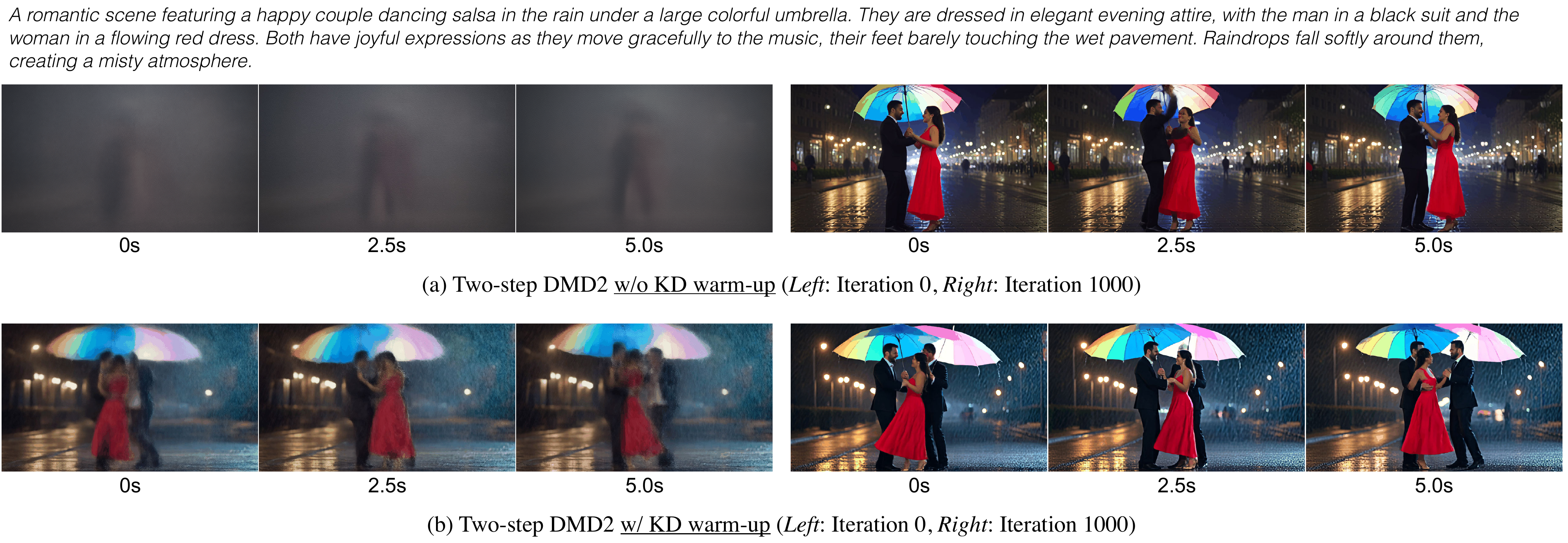}
    \vspace{-12pt}
    \caption{\small \textbf{Effect of KD initialization.} We compare the two-step DMD2 results of distilling Wan2.1 1.3B in two settings: (a) with and (b) without KD warm-up, where (left) ``Iteration 0'' means videos generated in the beginning of DMD2 training and (Right) ``Iteration 1000'' means videos generated after training DMD2 for 1k iterations. From left to right, we show the first, middle and the last frames in each video. We can see that the KD warm-up initially can generate better videos, but it also introduces coarse-grained artifacts. For instance, it generates an extra man besides a couple specified in the prompt. After training for 1k iterations, two-step DMD2 cannot remove these artifacts, leading to the worse generation quality than two-step DMD2 without KD warm-up. 
    }
    \label{fig:shift_ablation}
    \vspace{-5pt}
\end{figure*}

\subsection{DMD2}
\label{app:dmd2}

\paragraph{Initialization.}

Since we consider video diffusion models that are trained to approximate the instantaneous velocity $\vv$ in~\eqref{eq:outer_velocity}, we parametrize the student network as
\begin{equation}
\label{eq:student_param}
    \vg^\mathrm{student}_\theta(\vx_{t},t) = \vx_{t} - t \vv_\theta(\vx_{t},t)
\end{equation}
for our \dmd baseline, which initially approximates the conditional expectation $\mathbb{E}[\vx|\vx_{t}]$.

\paragraph{Multi-step distillation and inference.} Compared to the original multi-step DMD2~\citep{yin2024improved}, we did not use backward simulation for the outer loop for multi-step distillation and efficiently sample $\vx_{t_i}$ using real data.

Moreover, during inference, we use deterministic sampling from the conditional flow, \emph{i.e.},
\begin{equation}
\label{eq:det_sampling}
    \vx_{t_{i+1}} = \left(1 - \frac{t_{i+1}}{t_i}\right) \vx_{t_i} + \frac{t_{i+1}}{t_i} \vg^\mathrm{student}(\vx_{t_i},t_i),
\end{equation}
which samples from the correct distribution as long as $\vx \approx \mathrm{Student}(\vx_{t_i},t_i)$. Compared to the standard resampling scheme in DMD2, \emph{i.e.},
\begin{equation}
\label{eq:stoch_sampling}
    \vx_{t_{i+1}} = \left(1 - t_{i+1}\right) \vg^\mathrm{student}(\vx_{t_i},t_i) + t_{i+1} \vx_1,
\end{equation}
the noise $\vx_1$ of the conditional flow is inferred from $\vg^\mathrm{student}(\vx_{t_i},t_i)$ and $\vx_{t_i}$ in Eq.~\eqref{eq:det_sampling} and sampled independently in Eq.~\eqref{eq:stoch_sampling}.

For TMD, the multi-step inference is given in~\Cref{alg:dtm_generation}. In particular, the outer transitions are also deterministic, but additional independent noise $\vy_1\sim \mathcal{N}(0,I)$ is used for the inner flow. 

\paragraph{Teacher, fake score, and discriminator.}
While it would be possible to distill Wan2.1 1.3B using the 14B model as teacher in DMD2, we use the 1.3B model for fair comparisons. Moreover, we use (unpatchified) teacher features at different layers as inputs to the discriminator, which we found more stable than using fake score features. The discriminator is trained using an average minimax log-likelihood objective over separate heads for each teacher feature, where the teacher is evaluated at noisy inputs $(\hat{\vx}_t,t)$ (fake data as defined in~\Cref{sec:background}) or $(\vx_t,t)$ (real data). The generator loss is given as the average (non-saturating) negative log-likelihood. We initialize the fake score $\vg^{\mathrm{fake}}$ using the teacher parameters, parametrize it analogous to~\eqref{eq:student_param}, and use denoising score matching to train it on noisy fake data $(\hat{\vx}_t,t)$. Both the discriminator and fake score are trained for several iterations in between student updates. Finally, we can write our VSD objective in~\eqref{eq:vsd_objective} as
\begin{equation*}  \mathbb{E}_{t_i,\vx_{t_i},\tdmd,\hat{\vx}_{\tdmd}}\left[ \mathrm{sg}\Big(\frac{\vg^{\mathrm{fake}}(\hat{\vx}_{\tdmd},\tdmd) - \vg^{\mathrm{teacher}} (\hat{\vx}_{\tdmd},\tdmd)}{\|\vg^{\mathrm{fake}}(\hat{\vx}_{\tdmd},\tdmd) - \vg^{\mathrm{teacher}} (\hat{\vx}_{\tdmd},\tdmd)\|_1} \Big)^T \hat{\vx} \right],
\end{equation*}
where $\mathrm{sg}(\cdot)$ denotes the stop-gradient operation, and the teacher is also parametrized analogous to~Eq. \eqref{eq:student_param}, including CFG with the negative prompt given in~\Cref{tab:ext_prompts}.

\subsection{VBench evaluation}

We follow the official evaluation protocol in VBench~\cite{huang2023vbench} to test our method. During video sampling, we use standard VBench prompt lists to ensure fair comparisons among different methods. In particular, we rewrite the test prompts (\emph{i.e.}, prompt augmentation) using \textsc{Qwen/Qwen2.5-7B-Instruct}~\citep{yang2024qwen2}, following the prior work~\citep{huang2025self}. Similarly, almost all the baselines, including Wan2.1 base models, also enhance VBench prompts, making them longer and more descriptive without altering their original meaning. We then calculate scores for 16 Text-to-Video (T2V) evaluation dimensions, respectively, and summarize them into quality score, semantic score and overall score. 

In figures from the main text, we only show the short prompts due to the space limit. In Table~\ref{tab:ext_prompts}, we show the corresponding extended prompts in Figures~\ref{fig:example}-\ref{fig:prompt_adherance} that are actually passed to the model.

\begin{table*}[t]
\renewcommand{\arraystretch}{1.2} 
\setlength{\aboverulesep}{0pt}
\setlength{\belowrulesep}{0pt}
\vspace{-4pt}
\footnotesize
    \centering
    \begin{tabular}{p{0.11\linewidth} p{0.89\linewidth}}
    \toprule
    \rowcolor{lightgray!20}
    \emph{Figure 1} & \\
    Short prompt &  \emph{A fat rabbit wearing a purple robe walking through a fantasy landscape.}   \\
    Long prompt &  \emph{A plump, fluffy rabbit donning a voluminous purple robe walks gracefully through a vibrant fantasy landscape. The rabbit has large, expressive eyes and a gentle, curious expression. Its fur is soft and thick, and the robe drapes elegantly over its body. The landscape features rolling hills covered in lush green grass, colorful wildflowers, and towering magical trees with shimmering leaves. In the distance, there are sparkling waterfalls and mystical castles. The scene is bathed in warm, golden sunlight. Medium shot, focusing on the rabbit's walk through the picturesque environment.}  \\
    Short prompt &  \emph{A person drinking coffee in a cafe.} \\
    Long prompt &  \emph{A cozy, warm café setting with soft ambient lighting and wooden furnishings. A young adult, casually dressed in a sweater and jeans, sits at a small round table. They hold a steaming cup of coffee in their hand, taking a sip while looking pensively out the window. The café is moderately busy with other patrons engaged in conversations. The background showcases various coffee drinks and pastries displayed on a counter. The person’s expression is relaxed and content. Medium shot focusing on the person’s face and the coffee cup, capturing the intimate atmosphere of the café.}  \\

    \midrule    
    \rowcolor{lightgray!20}
    \emph{Figure 3} &  \\
    Short prompt &  \emph{A person is laughing.} \\
    Long prompt &  \emph{A joyful person is laughing heartily, with a broad smile and crinkled eyes, conveying pure happiness. They are standing upright with arms spread wide, as if embracing the world around them. The scene is set outdoors in a sunny park, surrounded by lush greenery and blooming flowers. The background includes a clear blue sky with fluffy clouds, adding to the cheerful atmosphere. Medium shot capturing the full body of the person, focusing on their animated facial expressions and gestures.}  \\

    Short prompt &  \emph{A steam train moving on a mountainside.} \\
    Long prompt &  \emph{A vintage steam locomotive chugging along a winding track on a mountainous terrain. The train is covered in soot, with steam billowing from its smokestack as it navigates the rugged landscape. The surrounding mountains are steep and lush, with patches of snow visible at higher elevations. The train cars sway gently as they follow the curving tracks, and the scenery outside the windows shows dense forests and rocky cliffs. The camera follows the train from a medium distance, capturing the train's movement and the dramatic backdrop.}  \\

    \midrule    
    \rowcolor{lightgray!20}
    \emph{Figure 4} &  \\
    Short prompt &  \emph{A boat sailing leisurely along the Seine River with the Eiffel Tower in background.} \\
    Long prompt &  \emph{A serene, picturesque scene of a small wooden boat gently gliding along the Seine River in Paris, France. The boat is rowed leisurely by a middle-aged man in a casual striped shirt and khaki pants, who rows smoothly with rhythmic strokes. The Eiffel Tower stands majestically in the background, partially visible through the misty morning air. The riverbank is lined with lush green trees and quaint buildings, reflecting off the calm waters. The overall atmosphere is peaceful and tranquil, capturing the essence of a lazy summer day. Wide shot, static camera.}  \\

    Short prompt &  \emph{An astronaut flying in space, zoom out.} \\
    Long prompt &  \emph{Astronaut floating in space with a helmet visor reflecting Earth below. The astronaut, wearing a full spacesuit with the American flag on the shoulder, is performing a spacewalk, arms extended as if in motion. The background shows the vastness of space with stars twinkling and Earth in the distance. The scene begins with a close-up of the astronaut and gradually zooms out to reveal the enormity of space surrounding them. Wide shot, showcasing the astronaut against the backdrop of the universe.}  \\
    
    \midrule\midrule    
    \rowcolor{lightgray!20}
    \emph{Negative prompt} & \\
    \multicolumn{2}{p{\linewidth}}{\emph{Bright tones, overexposed, static, blurred details, subtitles, style, works, paintings, images, static, overall gray, worst quality, low quality, JPEG compression residue, ugly, incomplete, extra fingers, poorly drawn hands, poorly drawn faces, deformed, disfigured, misshapen limbs, fused fingers, still picture, messy background, three legs, many people in the background, walking backwards}} \\
    \bottomrule
    \end{tabular}
    \vspace{-5pt}
    \caption{ \small Extended prompts used in the figures from the main text and negative prompt used for CFG (in teacher sampling, KD, and \dmd) and condition dropout (in TM-MF), taken from the official Wan repository.
    }
    \label{tab:ext_prompts}
    \vspace{-4pt}
\end{table*}

\subsection{Hyperparameters}

We provide an overview of our default hyperparameters in~\Cref{tab:hp}.

\vspace{-5pt}
\section{Additional experiments}
\label{app:add_exp}

\subsection{Quality-efficiency tradeoff}

\begin{figure*}
    \centering
    \includegraphics[width=0.9\linewidth]{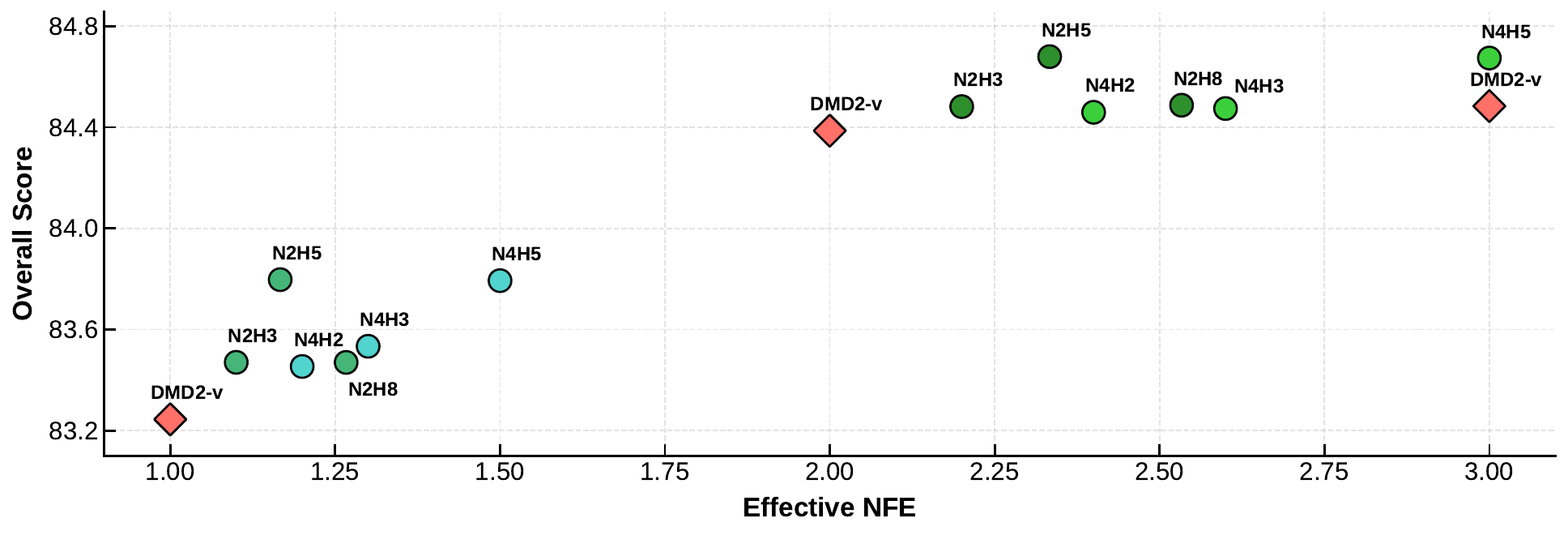}
    \vspace{-7pt}
    \caption{\textbf{Performance-efficiency tradeoff of TMD.} Extension of~\Cref{fig:tradeoff} to include the $M=1$ settings.}
    \label{fig:tradeoff_all}
    \vspace{-7pt}
\end{figure*}

Extending the experiments in~\Cref{sec:ablations},~\Cref{fig:tradeoff_all} shows TMD's performance-efficiency tradeoff when varying both the number of outer and inner steps $M$ and $N$ and the flow head layers $H$. It shows a general trend of achieving better performance with a larger effective NFE and TMD offers a more fine-grained control over this performance-efficiency tradeoff than DMD2-\emph{v}.

\begin{figure}[t]
    \centering
    \includegraphics[width=0.8\linewidth]{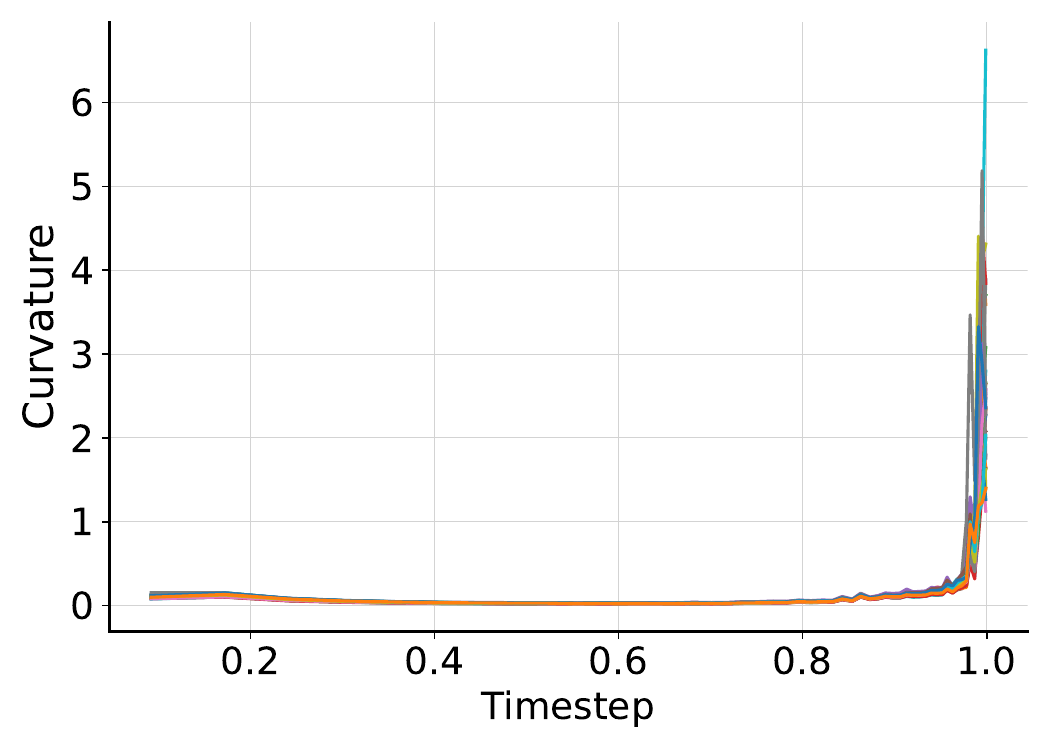}
    \vspace{-3pt}
    \caption{\textbf{Curvature of Wan trajectories.} Curvature of different sampling trajectories for the Wan2.1 1.3B model. Large trajectory curvatures are observed near $t = 1$ (\emph{i.e.}, the high noise regime).}
    \label{fig:curvature}
\end{figure}

\subsection{Curvature of Wan trajectories}

Inspired by~\citet{liu2023flow}, we define the curvature of Wan’s sampling trajectory at time $t$ as \begin{align*}
    C(\vv, t) = \Vert \vv(\vx_t, t) - (\vx - \vx_0)\Vert^2.
\end{align*}
Since the model is evaluated on a fixed time grid during sampling, we adopt a discretized version of the curvature in practice, \emph{i.e.}, \begin{align*}
    C(\vv, t_i) = \Big\Vert \tfrac{\tilde{\vx}_{t_i} - \tilde{\vx}_{t_{i-1}}}{t_i - t_{i-1}} - (\vx_1 - \vx)\Big\Vert^2,
\end{align*}
where $t_i$ denotes the timesteps from Wan’s default $50$-step sampling schedule and $\tilde{\vx}_{t_i}$ represents the corresponding intermediate samples along the trajectory.

As shown in~\Cref{fig:curvature}, the sampling trajectories of the Wan model exhibit extremely large curvature near $t = 1$ (\emph{i.e.}, the high noise regime), making it difficult for trajectory matching methods to learn the mapping along the ODE path. This also motivates our choice of larger $\gamma$ in the time-shifting function for $r_i$ in TM-MF (see~\Cref{app:tmmf}) and $t_{\text{student}}$ in \dmd (see~\Cref{sec:stage2}).

\subsection{Flow head conditioning}
\label{app:fuse_ablation}

Our main experiments use the gating mechanism explained in~\Cref{app:fuse} to condition the flow head on the main backbone's features $\vm = \vm(\vx,t)$. Another version is to concatenate $\vm$ and the patch embeddings of $\vy_s$ along the hidden dimension and then project to the original hidden size using a linear layer. To minimize the impact on the pretrained model, we initialized the weights of the linear layer as identity for the coordinates corresponding to $\vm$ and as a Gaussian with small standard deviation ($0.01$) for the remaining coordinates. In~\Cref{tab:fuse}, we see that TMD can achieve strong performance across fusion types. While convergence was more stable with the gating mechanism used for our main experiments,
concatenation is a strong alternative in terms of final performance.

\begin{table}[H]
\footnotesize
    \centering
    \setlength{\tabcolsep}{2pt}
    \begin{tabular}{l l c c c}
    \toprule
    \multirow{2}{*}{Setting} & \multirow{2}{*}{Fusion type} & Overall  & Quality  & Semantic  \\
    & & score  & score  & score  \\
    \midrule    
    \multirow{2}{*}{N2H5}  & Gated (see~\Cref{app:fuse}) & 84.68 & 85.71 & 80.55 \\
      & Concat & \textbf{84.76} & \textbf{85.77} & \textbf{80.71} \\
    \midrule
    \multirow{2}{*}{N4H5}  & Gated (see~\Cref{app:fuse}) & \textbf{84.67} & 85.72 & \textbf{80.47} \\
    & Concat & 84.66 & \textbf{85.79} & 80.14 \\
    \bottomrule
    \end{tabular}
    \vspace{-5pt}
    \caption{Impact of the fusion type on the final performance when distilling Wan2.1 1.3B with $M=2$.}
    \label{tab:fuse}
    \vspace{-3pt}
\end{table}

\vspace{-7pt}
\subsection{Inner flow targets}
\label{app:inner_flow_targets}

In principle, the auxiliary latent variable $\vy$ in the inner flow can be chosen arbitrarily as long as $\vx_{t_{i-1}}$ is easy to sample given $\vy$ and $\vx_{t_{i}}$.
While we chose the DTM formulation $\vy \coloneqq \vx_1 - \vx$ in~\eqref{eq:dtm} for our main experiments, we also experimented with other versions. While other formulations did not improve the overall performance, they could still achieve competitive performance. 
For instance,~\Cref{tab:target} provides results for the simple choice $\vy\coloneqq \vx$. Similar to~\eqref{eq:tmmf_precond} and~\eqref{eq:student_param}, we parametrize the average velocity as
\begin{equation*}
    \vu_\theta(\vy_{s}, s, r; \vm) \coloneqq \frac{\vy_{s} - (\vx_{t_i} - t_i\mathrm{head}_\theta(\vy_{s}, s, r; \vm))}{s}.
\end{equation*}
where $\vm=\vm_\theta(x_{t_i},t_i)$ denote the main features.

\subsection{Impact of recurrence in flow head}

\begin{figure*}[t!]
    \centering
    \includegraphics[width=\linewidth]{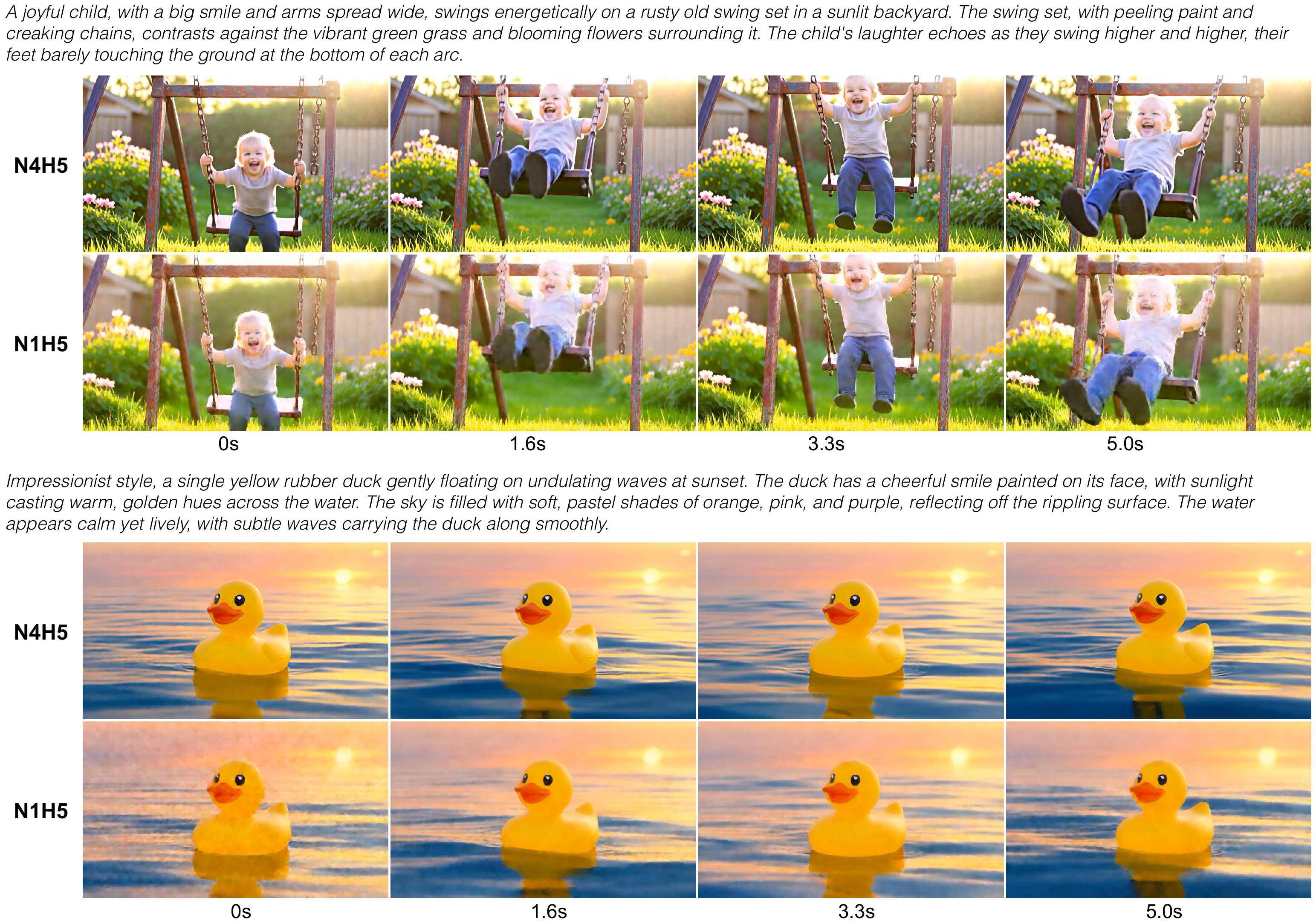}
    \vspace{-12pt}
    \caption{\small \textbf{Impact of flow head recurrence.} We show \underline{the impact of recurrence} in the flow head by setting the number of flow head steps to 1 \underline{only at inference} (\emph{i.e.}, N1H5) when distilling Wan2.1 1.3B with $M=2$ and the N4H5 setting for flow head (\emph{i.e.}, 4 denoising steps and 5 DiT blocks in flow head). We observe that the videos generated without recurrence (marked by N1H5) are of much lower quality (\emph{e.g.}, more artifacts and blurriness) than ones with recurrence (marked by N4H5), implying the importance of the fine-grained iterative refinement on our method. 
    }
    \label{fig:flow_steps_ablation}
    \vspace{-5pt}
\end{figure*}

To evaluate the effect of recurrence in the flow head, we manually restrict the inner flow to a single step at inference (N1H5) and compare the results with the standard recurrent setting (N4H5). As shown in Figure~\ref{fig:flow_steps_ablation}, when distilling Wan2.1 1.3B into a two-step generator, the non-recurrent variant produces noticeably lower-quality videos, exhibiting stronger artifacts and blurriness. This highlights the importance of iterative refinement within the flow head for high-fidelity generation.

\begin{table}[H]
\footnotesize
    \centering
    \begin{tabular}{l l c c c}
    \toprule
    \multirow{2}{*}{Setting} & \multirow{2}{*}{Target $\vy$} & Overall  & Quality  & Semantic  \\
    & & score  & score  & score  \\
    \midrule    
    \multirow{2}{*}{N2H5}  & $\vx_1 - \vx$ (DTM) & \textbf{84.68} & \textbf{85.71} & \textbf{80.55} \\
      & $\vx$ & 84.18 & 85.15 & 80.33 \\
    \midrule
    \multirow{2}{*}{N4H5}  & $\vx_1 - \vx$ (DTM) & \textbf{84.67} & \textbf{85.72} & 80.47 \\
    & $\vx$ & 84.44 & 85.36 & \textbf{80.77} \\
    \bottomrule
    \end{tabular}
    \vspace{-5pt}
    \caption{Impact of the inner flow target when distilling Wan2.1 1.3B with $M=2$.}
    \label{tab:target}
    \vspace{-3pt}
\end{table}

\subsection{DMD2 comparisons}
In~\Cref{sec:stage2}, we observed that KD pretaining is only helpful for one-step distillation (see~\Cref{tab:kd} and~\Cref{fig:kd_ablation}) and that timestep shifting of $t_{\mathrm{dmd}}$ and $t_{\mathrm{student}}$ is crucial for good performance for one- and two-step \dmd (see~\Cref{tab:t_shift} and~\Cref{fig:shift_ablation}).
In this section, we provide additional ablations on our \dmd baseline when using $\ge 3$ steps. While the timestep shifting for $t_\text{dmd}$ and $t_\text{student}$ remain important, we observed that slightly lower shift values for $t_\text{student}$ can attain better scores when using more steps. In particular, for $4$ steps, we picked a shift value of $\gamma=5$ for $t_{\mathrm{student}}$, outperforming the DMD2 baseline in~\cite{zheng2025rcm}; see~\Cref{tab:dmd2}.

\begin{table}[H]
\footnotesize
    \centering
    \setlength{\tabcolsep}{3pt}
    \begin{tabular}{l c c c c c}
    \toprule
    \multirow{2}{*}{Method} & \multirow{2}{*}{shift $t_{\text{student}}$} & \multirow{2}{*}{NFE} & Overall & Quality  & Semantic  \\
    & & & score  & score  & score  \\
    \midrule
    DMD2~\citep{zheng2025rcm} & & 4 & {84.56} & {85.58} & \textbf{80.50} \\   
    \dmd & 10 & 4 & 84.53 & 85.95 & 78.84 \\
    \dmd & 5 & 4 & \textbf{84.60} & \textbf{86.03} & 79.87 \\
    \midrule 
    \dmd & 10 & 3 & \textbf{84.48} & \textbf{85.71} & 79.58 \\
    \dmd & 5 & 3 & 84.47 & 85.55 & \textbf{80.15} \\
    \bottomrule
    \end{tabular}
    \vspace{-5pt}
    \caption{VBench results of the DMD2 version in~\citep{zheng2025rcm} and our \dmd for $3$ and $4$ steps with different timestep shifts for $t_{\text{student}}$. 
    }
    \vspace{-3pt}
    \label{tab:dmd2}
\end{table}

\subsection{Inference time}

In Table~\Cref{tab:latency_breakdown}, we measure the inference time of our proposed decoupled architecture to confirm that the flow head is lightweight compared to the backbone. Moreover, the gray columns validate the usage of \emph{effective} NFE for evaluating efficiency, since the increase in the per-step inference time closely matches the effective NFE.

\begin{table}[h]
\centering
\footnotesize
\setlength{\tabcolsep}{4pt}
\begin{tabular}{lcccccc}
\toprule
Setting & Backbone & Head & Fusion & Total & \textcolor{gray}{Increase} & \textcolor{gray}{NFE} \\
\midrule
N4H5 (1.3B) & 783 & 156 & 4.8 & 1426 & \textcolor{gray}{1.51x} & \textcolor{gray}{1.5} \\
N2H5 (1.3B) & 781 & 155 & 4.7 & 1100 & \textcolor{gray}{1.17x} & \textcolor{gray}{1.17} \\
N4H5 (14B)  & 4191 & 595 & 29  & 6687 & \textcolor{gray}{1.4x} & \textcolor{gray}{1.38} \\
\bottomrule
\end{tabular}
\vspace{-5pt}
\caption{Inference time (in milliseconds) of different architecture components with $M=1$ and batch-size $1$ on a single H100 GPU. The increase in inference time (second-to-last column) is measured relative to a standard forward pass of the teacher model and closely matches the effective NFE (last column).}
\label{tab:latency_breakdown}
\vspace{-1.25em}
\end{table}

\subsection{Training iterations}

We note that TMD is training-efficient compared to other baselines. For instance, for Wan2.1 14B and $M=2$ we only require $3$k iterations of lightweight TM-MF pretraining and $1$k iterations of distribution matching, whereas rCM uses $10$k iterations (with VSD and sCM losses)~\cite{zheng2025rcm}. For Wan2.1 1.3B, we show in~\Cref{tab:iter} that even with significantly reduced number of TM-MF iterations, TMD still outperforms rCM and DMD2-\emph{v}.

\begin{table}[h ]
\footnotesize
\setlength{\columnsep}{0pt}
\label{tab:cost}
\begin{tabular}{lccccc}
\toprule
\multirow{2}{*}{Method} & TM-MF & Distill &  Overall  & Quality  & Semantic  \\
& (iters) & (iters) & score  & score  & score  \\
\midrule
rCM~\cite{zheng2025rcm} & - & \phantom{$<$}10k & 84.09  & 84.90 & \textbf{80.86} \\
DMD2-\emph{v} & - & $<$6k & 84.39  & \underline{85.65} & 79.32 \\
TMD-N2H5 & 0.5k & $<$6k & 84.52  &  \underline{85.65} & 79.99 \\
TMD-N2H5 & 1k & $<$6k & \underline{84.59}  & 85.57 & \underline{80.68} \\
TMD-N2H5 & 3k & $<$6k & \textbf{84.68} & \textbf{85.71} & 80.55 \\
\bottomrule
\end{tabular}
\vspace{-5pt}
\caption{Comparison of VBench scores for distilling Wan2.1 1.3B using TMD ($M=2$) with a varying number of TM-MF pretraining iterations against two-step baselines.}
\label{tab:iter}
\end{table}

\subsection{More visual comparison results}

\begin{figure*}[t!]
    \vspace{-5pt}
    \centering
    \hspace{0.005\linewidth}%
    \begin{minipage}{0.87\linewidth}
    {\fontfamily{DejaVuSans-TLF}\selectfont\tiny\emph{
    A person enjoying a juicy burger, with a satisfied smile on their face. They are seated at a casual dining table, surrounded by napkins and a drink. The burger is topped with lettuce, tomato, and cheese, and the person is taking a bite, showcasing the delicious layers inside. The scene has a warm, inviting atmosphere with soft lighting and a cozy background. Medium close-up shot focusing on the person's hand holding the burger and their facial expressions as they savor each bite.}\par}\vspace{-0.05em}
    \end{minipage}%
    \hspace{0.125\linewidth}
    \includegraphics[width=0.9\linewidth,trim={0 0 0 5pt}, clip]{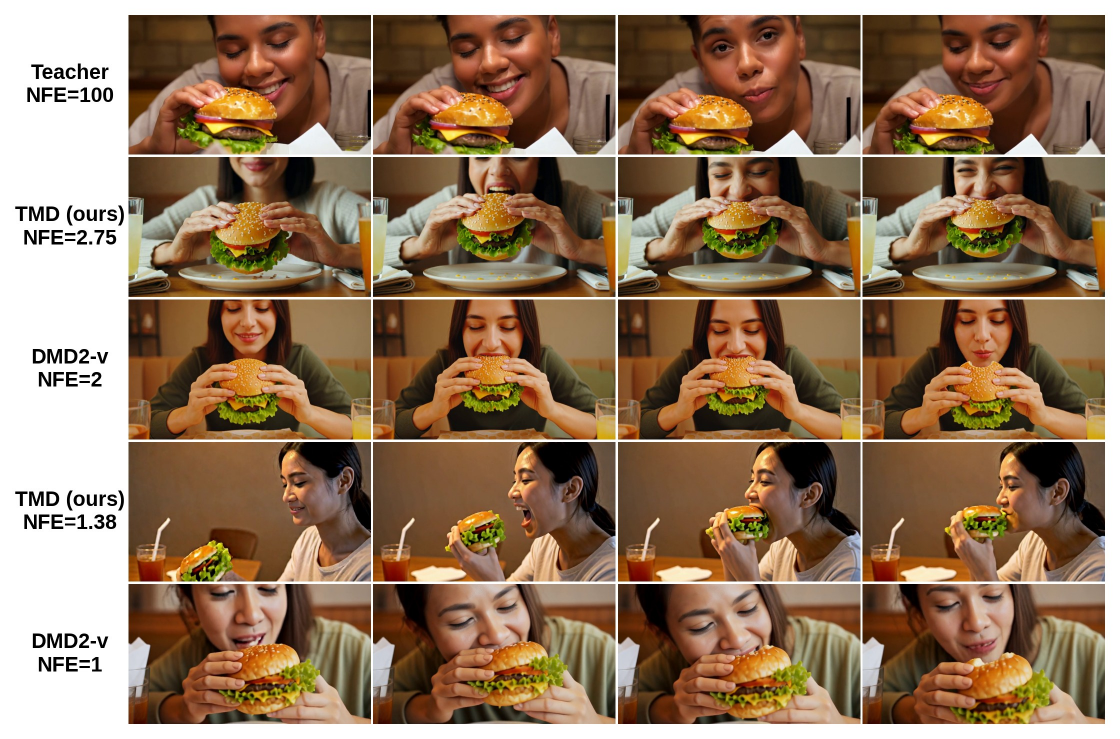}
    \hspace{0.005\linewidth}%
    \begin{minipage}{0.87\linewidth}
    {\fontfamily{DejaVuSans-TLF}\selectfont\tiny\emph{
    A person is skydiving from a plane, descending towards the ground. They are mid-air, arms spread wide, with a parachute deployed and open, ensuring a smooth descent. The skydiver is wearing a jumpsuit and helmet, with a determined and exhilarated expression on their face. The background shows a clear blue sky with fluffy clouds and the landscape below stretching out, including patches of green fields and distant mountains. The scene captures the moment just after exiting the plane, with the parachute fully inflated, showcasing the thrill and freedom of skydiving. Mid-shot, focusing on the skydiver against the expansive sky.}\par}\vspace{-0.05em}
    \end{minipage}%
    \hspace{0.125\linewidth}
    \includegraphics[width=0.9\linewidth,trim={0 0 0 5pt}, clip]{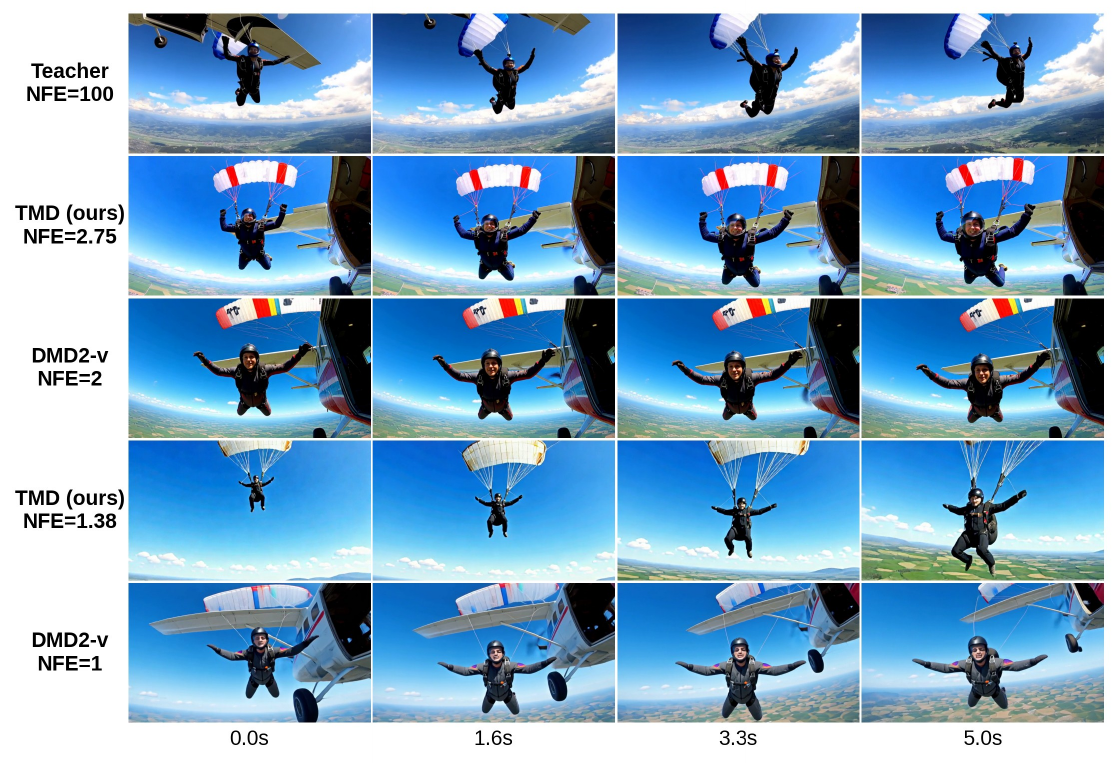}
    \vspace{-10pt}
    \caption{\textbf{Visual comparison on Wan2.1 14B.} We compare the outputs of the teacher, TMD, and \dmd on exemplary prompts.}
    \label{fig:viz_14b_1}
    \vspace{-5pt}
\end{figure*}

\begin{figure*}[t!]
    \vspace{-5pt}
    \centering
    \hspace{0.005\linewidth}%
    \begin{minipage}{0.87\linewidth}
    {\fontfamily{DejaVuSans-TLF}\selectfont\tiny\emph{
    A serene African savanna scene with tall grasses and scattered trees. A tall giraffe, with distinctive brown spots on its creamy white coat, bends its long neck gracefully to drink from a calm river. The giraffe's gaze is focused intently on the water as it lowers its head, revealing its long eyelashes and gentle expression. The river reflects the golden hues of the late afternoon sun, casting a warm glow over the scene. The background shows the vast expanse of the savanna with distant hills. The video is a medium close-up, capturing the giraffe's elegant movement and the tranquil environment.}\par}\vspace{-0.05em}
    \end{minipage}%
    \hspace{0.125\linewidth}
    \includegraphics[width=0.9\linewidth,trim={0 0 0 5pt}, clip]{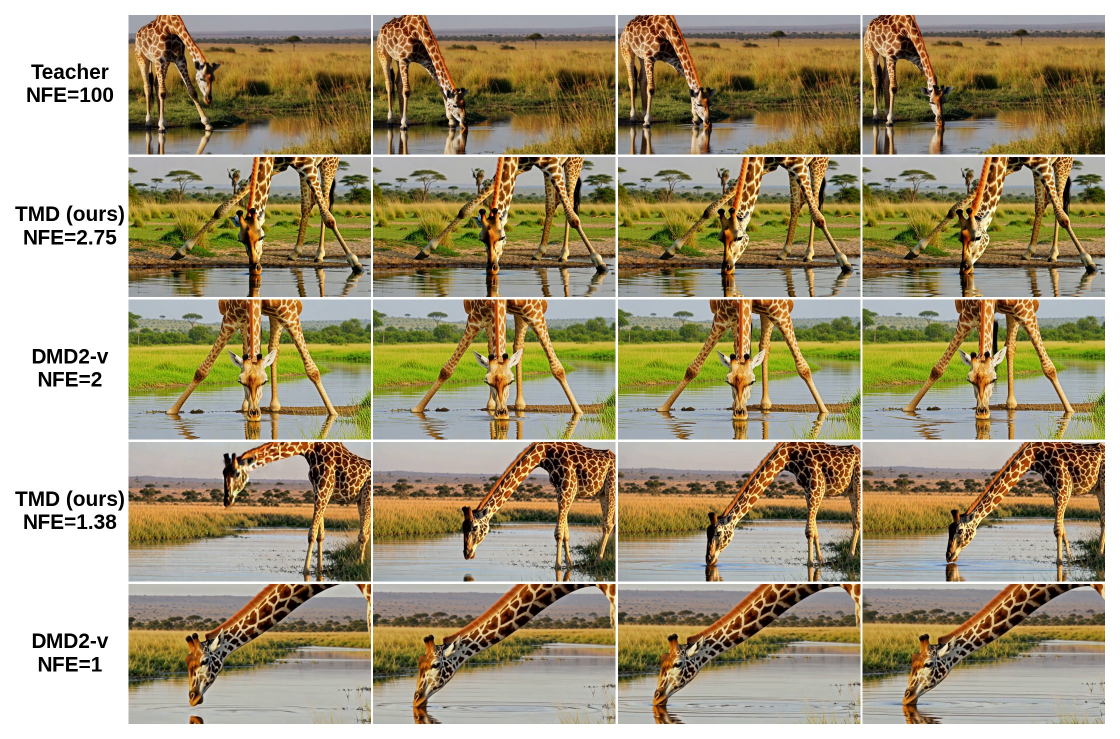}
    \hspace{0.005\linewidth}%
    \begin{minipage}{0.87\linewidth}
    {\fontfamily{DejaVuSans-TLF}\selectfont\tiny\emph{
    A large great white shark is swimming gracefully through the vast, deep blue ocean. Its sleek, muscular body cuts through the water as it propels forward with powerful tail strokes. The shark's dorsal fin slices through the surface, while smaller fish dart around it. The camera begins at a wide shot of the shark and the surrounding ocean, then smoothly zooms in to focus closely on the shark's sharp teeth and piercing eyes. The scene is filled with sunlight filtering through the water, creating a dynamic interplay of light and shadow. Close-up underwater perspective.}\par}\vspace{-0.05em}
    \end{minipage}%
    \hspace{0.125\linewidth}
    \includegraphics[width=0.9\linewidth,trim={0 0 0 5pt}, clip]{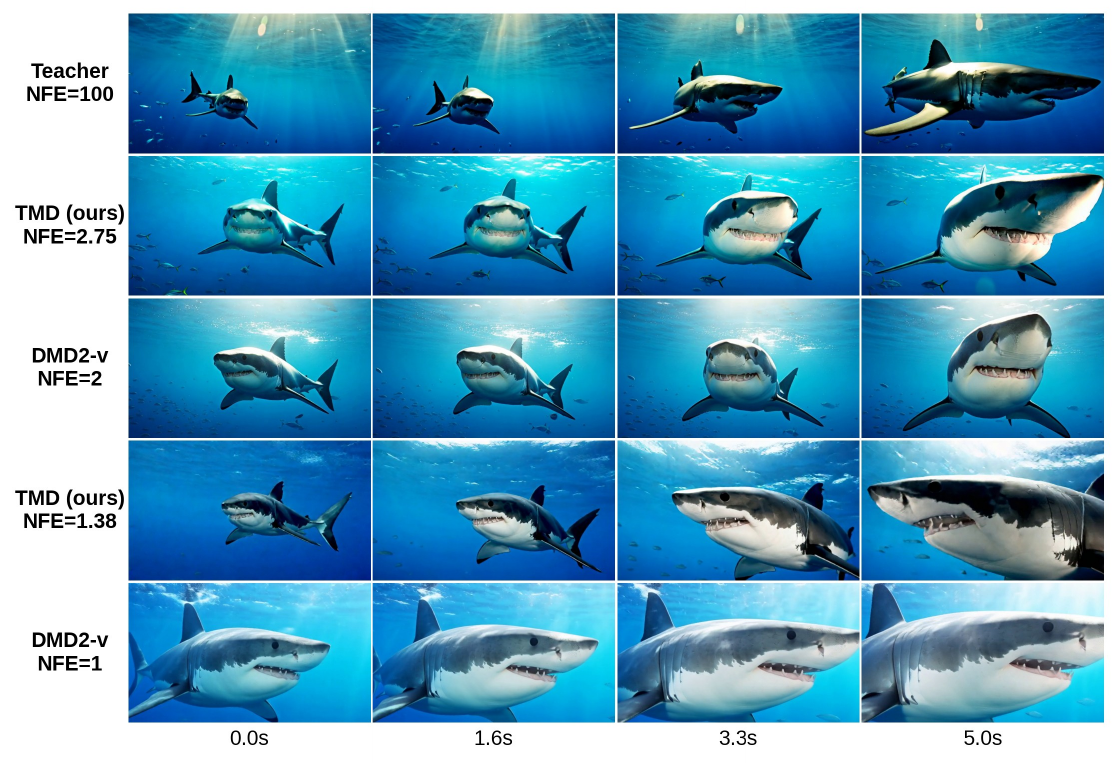}
    \vspace{-10pt}
    \caption{\textbf{Visual comparison on Wan2.1 14B.} We compare the outputs of the teacher, TMD, and \dmd on exemplary prompts.}
    \label{fig:viz_14b_2}
    \vspace{-5pt}
\end{figure*}

\begin{figure*}[t!]
    \vspace{-5pt}
    \centering
    \hspace{0.005\linewidth}%
    \begin{minipage}{0.87\linewidth}
    {\fontfamily{DejaVuSans-TLF}\selectfont\tiny\emph{
    A happy, playful Corgi running and jumping in a park during sunset, captured in black and white. The Corgi has a friendly face with floppy ears and a wagging tail as it moves through the grassy area. The sky behind the dog shows soft gradients of orange and pink fading into shades of gray and black. The park includes a few trees and benches in the background, adding depth to the scene. The Corgi is in motion, emphasizing its joyful playfulness. Medium close-up shot, focusing on the Corgi's expressive face and body language.}\par}\vspace{-0.05em}
    \end{minipage}%
    \hspace{0.125\linewidth}
    \includegraphics[width=0.9\linewidth,trim={0 0 0 5pt}, clip]{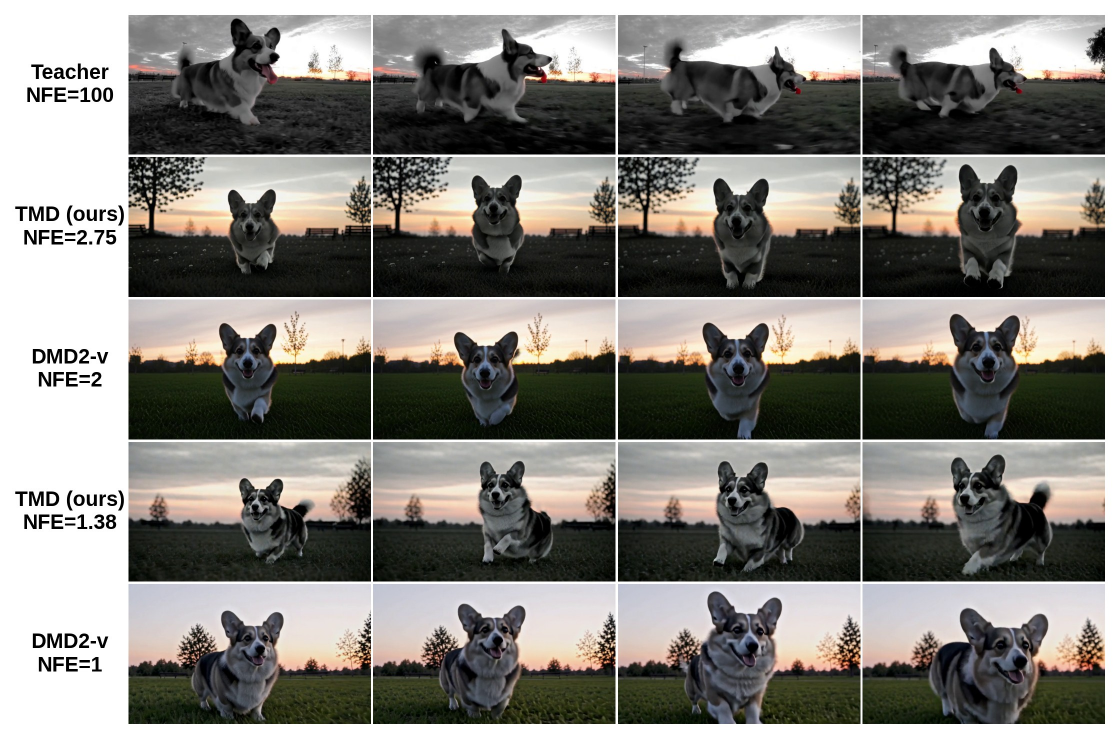}
    \hspace{0.005\linewidth}%
    \begin{minipage}{0.87\linewidth}
    {\fontfamily{DejaVuSans-TLF}\selectfont\tiny\emph{
    A stormtrooper from the Star Wars universe, clad in pristine white armor with a black helmet, is meticulously vacuuming a sandy beach. He bends down slightly, moving the vacuum cleaner back and forth across the sand with purposeful motions. His gloved hand firmly grips the handle of the vacuum as he navigates around rocks and debris. The sun sets behind him, casting long shadows and giving the scene a dramatic, golden glow. The background shows crashing waves and seagulls flying overhead. Medium close-up shot, focusing on the stormtrooper's actions and the sweeping motion of the vacuum.}\par}\vspace{-0.05em}
    \end{minipage}%
    \hspace{0.125\linewidth}
    \includegraphics[width=0.9\linewidth,trim={0 0 0 5pt}, clip]{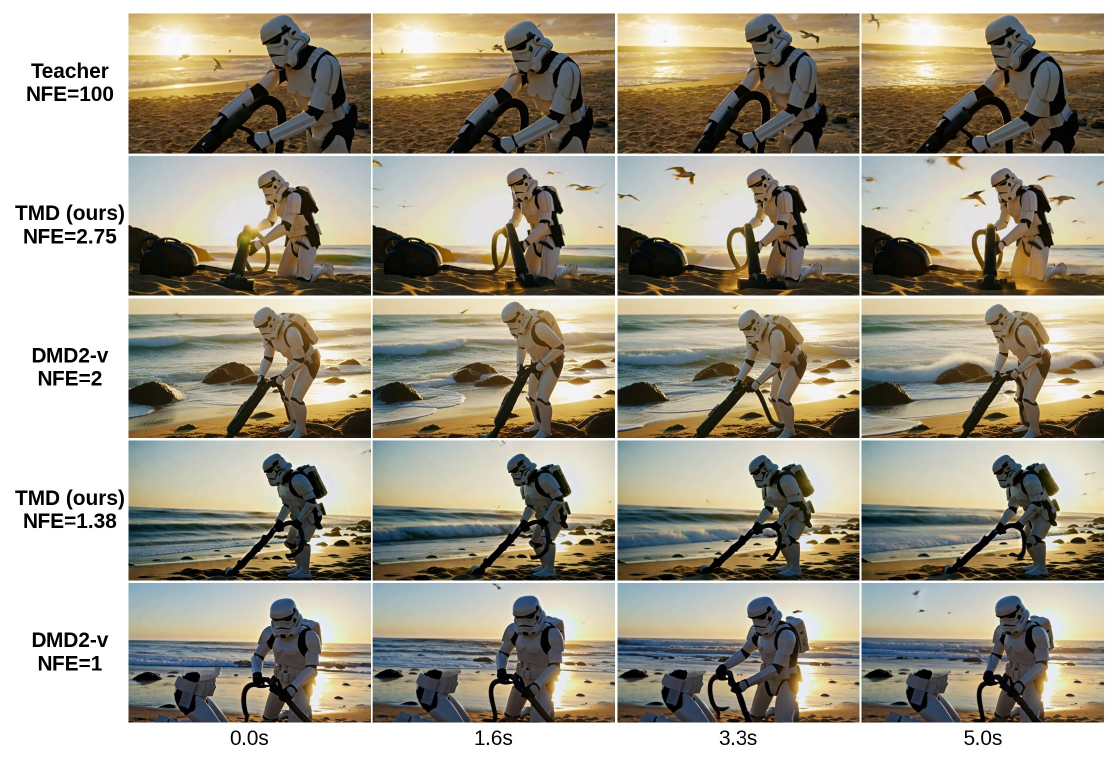}
    \vspace{-10pt}
    \caption{\textbf{Visual comparison on Wan2.1 14B.} We compare the outputs of the teacher, TMD, and \dmd on exemplary prompts.}
    \label{fig:viz_14b_3}
    \vspace{-5pt}
\end{figure*}

\begin{figure*}[t!]
    \vspace{-5pt}
    \centering
    \hspace{0.005\linewidth}%
    \begin{minipage}{0.87\linewidth}
    {\fontfamily{DejaVuSans-TLF}\selectfont\tiny\emph{
    A person is roller skating in an urban park, moving smoothly across the paved path. They wear a black helmet with a visor and knee pads, elbow pads, and wrist guards for safety. The skater has medium-length brown hair tied back in a ponytail and wears a bright yellow shirt with a graphic design and black shorts. They maintain a slight crouch posture as they glide, their feet making fluid movements, propelling them forward effortlessly. The background shows other park-goers walking dogs and children playing, adding a lively atmosphere. Medium shot focusing on the skater from a side angle, capturing the motion and environment.}\par}\vspace{-0.05em}
    \end{minipage}%
    \hspace{0.125\linewidth}
    \includegraphics[width=0.9\linewidth,trim={0 0 0 5pt}, clip]{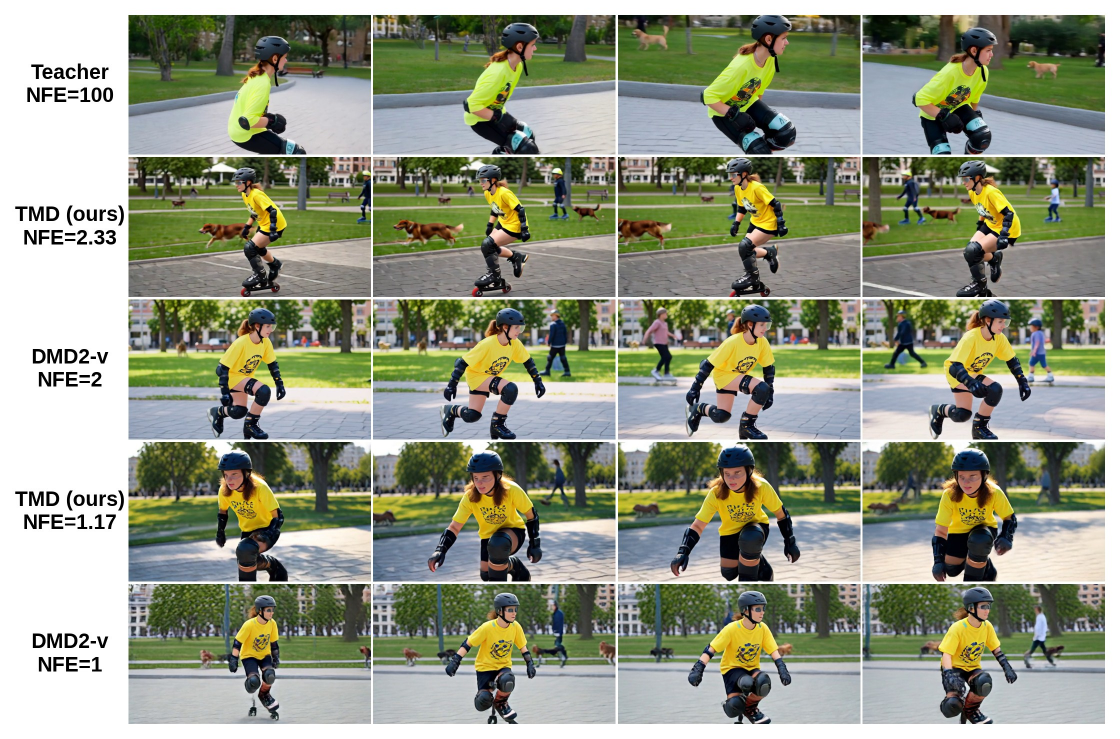}
    \hspace{0.005\linewidth}%
    \begin{minipage}{0.87\linewidth}
    {\fontfamily{DejaVuSans-TLF}\selectfont\tiny\emph{
    A koala bear playing a grand piano in a lush, dense forest. The koala has soft, grey fur and large, round ears. It sits upright on the piano bench, paws delicately placed on the keys, creating gentle melodies. The forest background is filled with tall eucalyptus trees, dappled sunlight filtering through the leaves, and a carpet of green moss beneath the piano. The scene is calm and serene, with the koala focused intently on its performance. Medium close-up shot, capturing the koala and part of the forest surroundings.}\par}\vspace{-0.05em}
    \end{minipage}%
    \hspace{0.125\linewidth}
    \includegraphics[width=0.9\linewidth,trim={0 0 0 5pt}, clip]{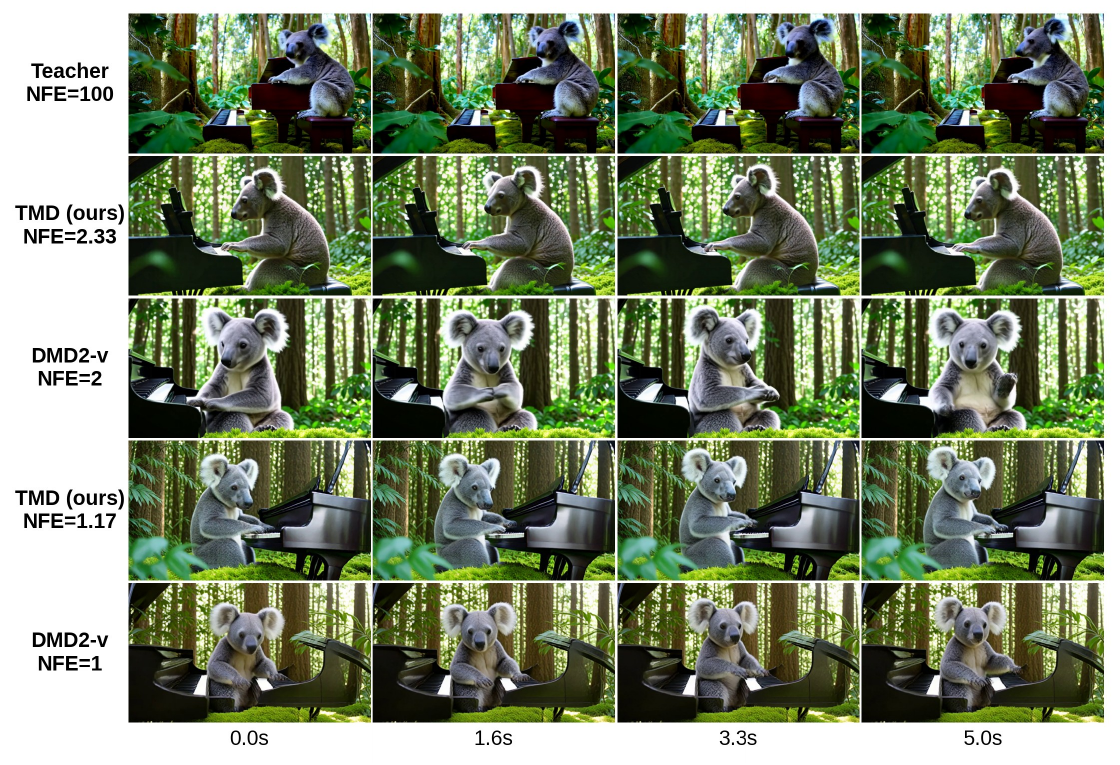}
    \vspace{-10pt}
    \caption{\textbf{Visual comparison on Wan2.1 1.3B.} We compare the outputs of the teacher, TMD, and \dmd on exemplary prompts.}
    \label{fig:viz_1_3b_1}
    \vspace{-5pt}
\end{figure*}

\begin{figure*}[t!]
    \vspace{-5pt}
    \centering
    \hspace{0.005\linewidth}%
    \begin{minipage}{0.87\linewidth}
    {\fontfamily{DejaVuSans-TLF}\selectfont\tiny\emph{
    A large, hairy Bigfoot creature walking through a heavy snowstorm. The Bigfoot stands at least eight feet tall, covered in shaggy brown fur, with muscular limbs and a stooped posture. Snowflakes swirl around him as he moves slowly and deliberately through the dense forest, his feet sinking slightly into the deep snow. The landscape is bleak and desolate, with bare trees and thick snowdrifts. The atmosphere is eerie and quiet, with only the sound of crunching snow and howling wind. The scene is captured in a mid-shot, focusing on the Bigfoot’s powerful form as he trudges through the storm.}\par}\vspace{-0.05em}
    \end{minipage}%
    \hspace{0.125\linewidth}
    \includegraphics[width=0.9\linewidth,trim={0 0 0 5pt}, clip]{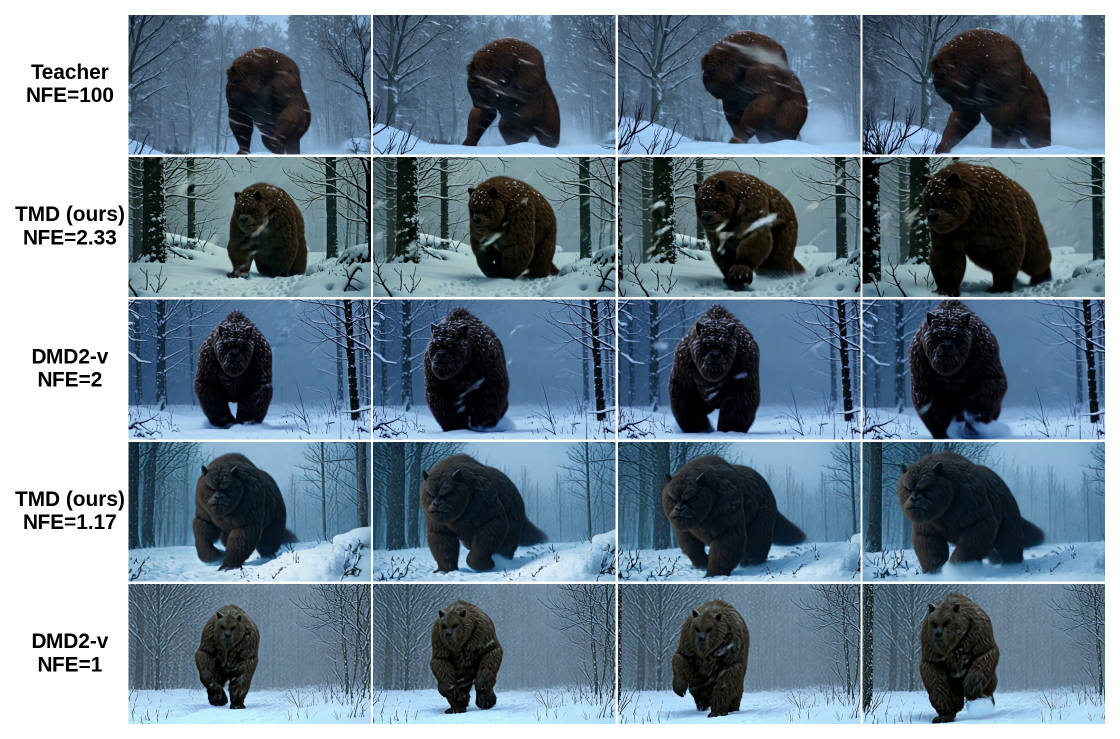}
    \hspace{0.005\linewidth}%
    \begin{minipage}{0.87\linewidth}
    {\fontfamily{DejaVuSans-TLF}\selectfont\tiny\emph{
    An animated scene of a panda drinking coffee in a cozy café in Paris. The panda is sitting at a small table with a steaming cup of coffee, holding a spoon delicately. The café has vintage decor with wooden furniture, soft lighting, and a few other patrons in the background. The panda has a relaxed and content expression, sipping the coffee slowly. The atmosphere is warm and inviting, with the soft hum of conversation in the background. Medium shot focusing on the panda and the coffee cup.}\par}\vspace{-0.05em}
    \end{minipage}%
    \hspace{0.125\linewidth}
    \includegraphics[width=0.9\linewidth,trim={0 0 0 5pt}, clip]{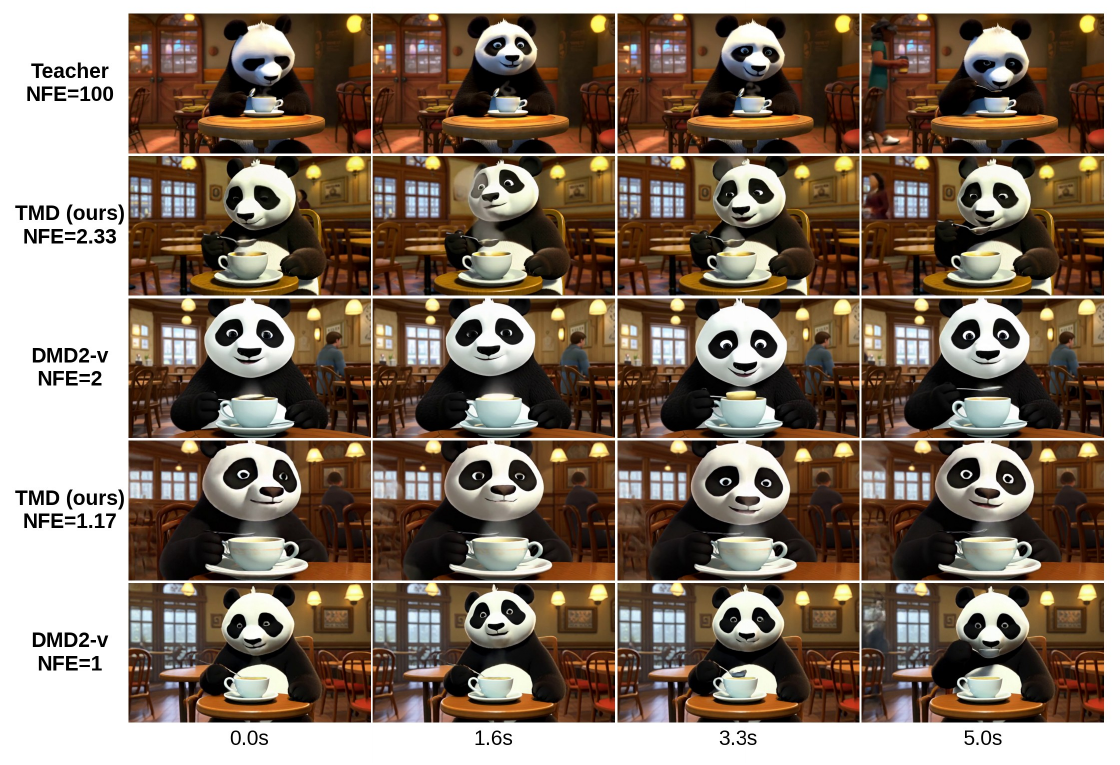}
    \vspace{-10pt}
    \caption{\textbf{Visual comparison on Wan2.1 1.3B.} We compare the outputs of the teacher, TMD, and \dmd on exemplary prompts.}
    \label{fig:viz_1_3b_2}
    \vspace{-5pt}
\end{figure*}

\begin{figure*}[t!]
    \vspace{-5pt}
    \centering
    \hspace{0.005\linewidth}%
    \begin{minipage}{0.87\linewidth}
    {\fontfamily{DejaVuSans-TLF}\selectfont\tiny\emph{
    A dramatic and intense scene featuring an erupting volcano. The volcano is spewing lava and ash into the air, creating a vivid orange glow against a dark night sky filled with billowing smoke clouds. The ground trembles as molten rock flows down the sides of the volcano, lighting up the surrounding landscape. In the foreground, a few scattered trees and rocks are illuminated by the fiery eruption. The camera remains fixed on the volcano, capturing the powerful motion and scale of the event. Nighttime, wide shot.}\par}\vspace{-0.05em}
    \end{minipage}%
    \hspace{0.125\linewidth}
    \includegraphics[width=0.9\linewidth,trim={0 0 0 5pt}, clip]{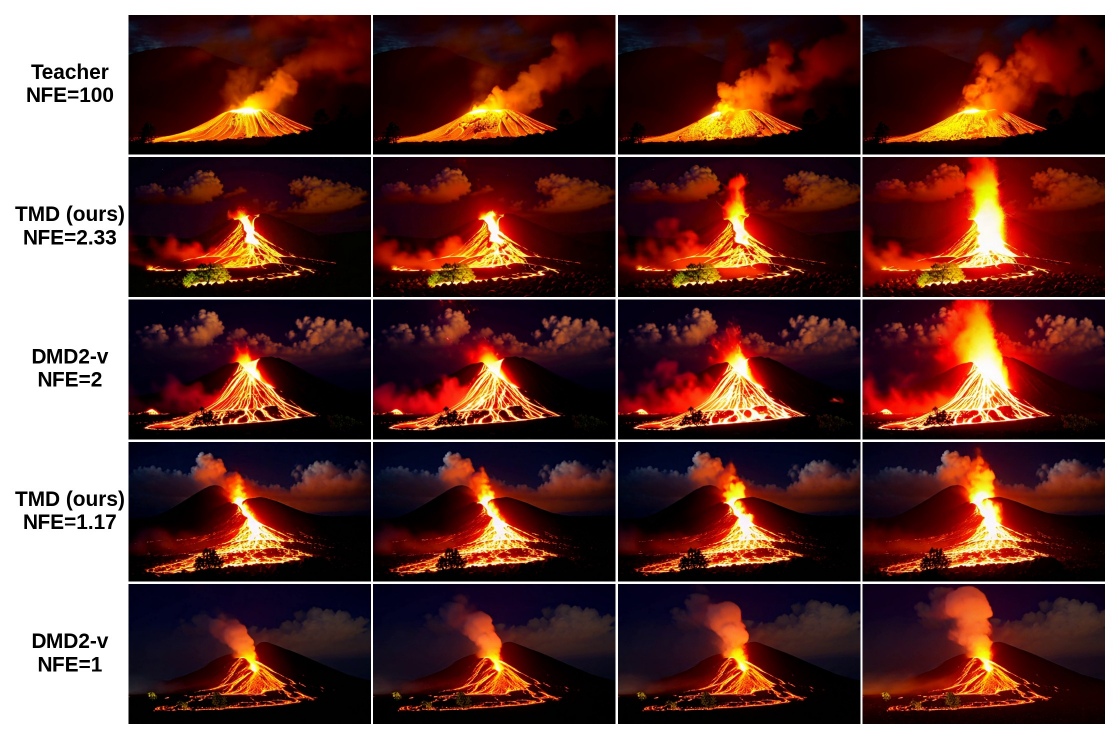}
    \hspace{0.005\linewidth}%
    \begin{minipage}{0.87\linewidth}
    {\fontfamily{DejaVuSans-TLF}\selectfont\tiny\emph{
    A realistic classroom scene set during a typical school day. The classroom has rows of desks facing a chalkboard at the front. Students are engaged in various activities; some are reading books, others are writing in notebooks, and a few are quietly talking. The teacher stands at the front of the class, holding a book and addressing the students. The room is well-lit with sunlight streaming in from large windows, casting soft shadows across the desks. The walls are adorned with educational posters and motivational quotes. Medium shot capturing the full classroom environment.}\par}\vspace{-0.05em}
    \end{minipage}%
    \hspace{0.125\linewidth}
    \includegraphics[width=0.9\linewidth,trim={0 0 0 5pt}, clip]{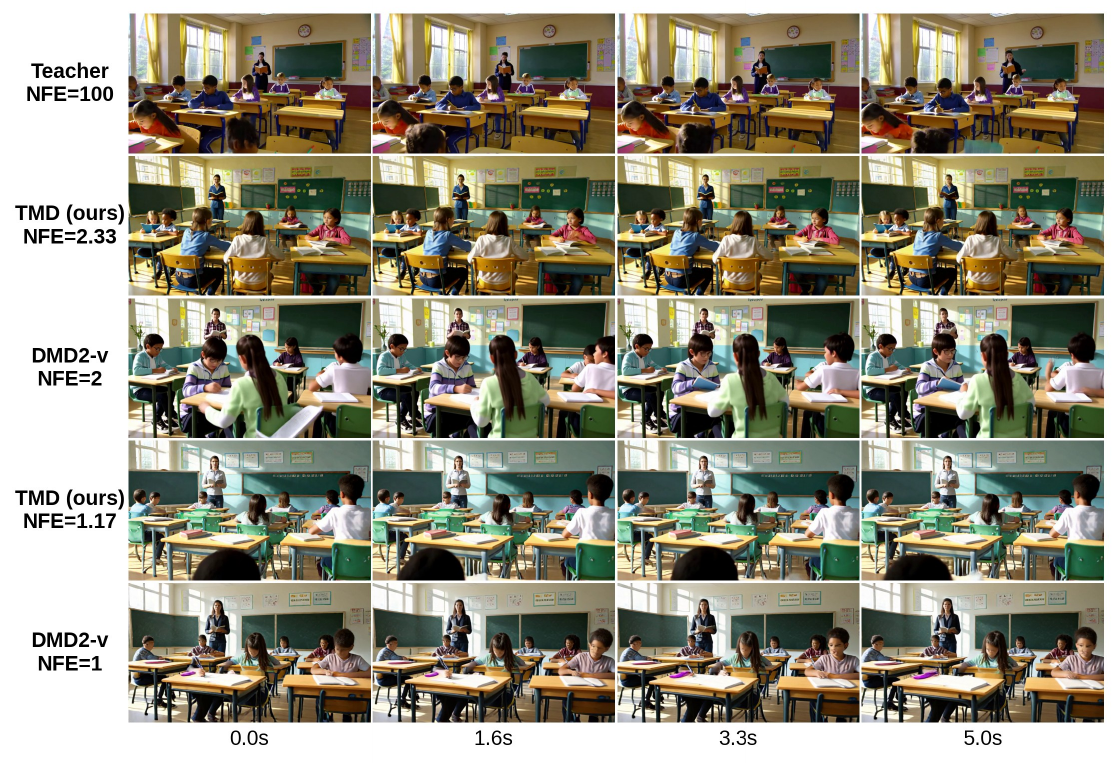}
    \vspace{-10pt}
    \caption{\textbf{Visual comparison on Wan2.1 1.3B.} We compare the outputs of the teacher, TMD, and \dmd on exemplary prompts.}
    \label{fig:viz_1_3b_3}
    \vspace{-5pt}
\end{figure*}

We provide further visual comparisons between the 50-step teacher models with classifier-free guidance (CFG), \dmd, and TMD in~\Cref{fig:viz_14b_1,fig:viz_14b_2,fig:viz_14b_3,fig:viz_1_3b_1,fig:viz_1_3b_2,fig:viz_1_3b_3}.

\begin{figure*}[t!]
    \centering
    \includegraphics[width=0.9\linewidth]{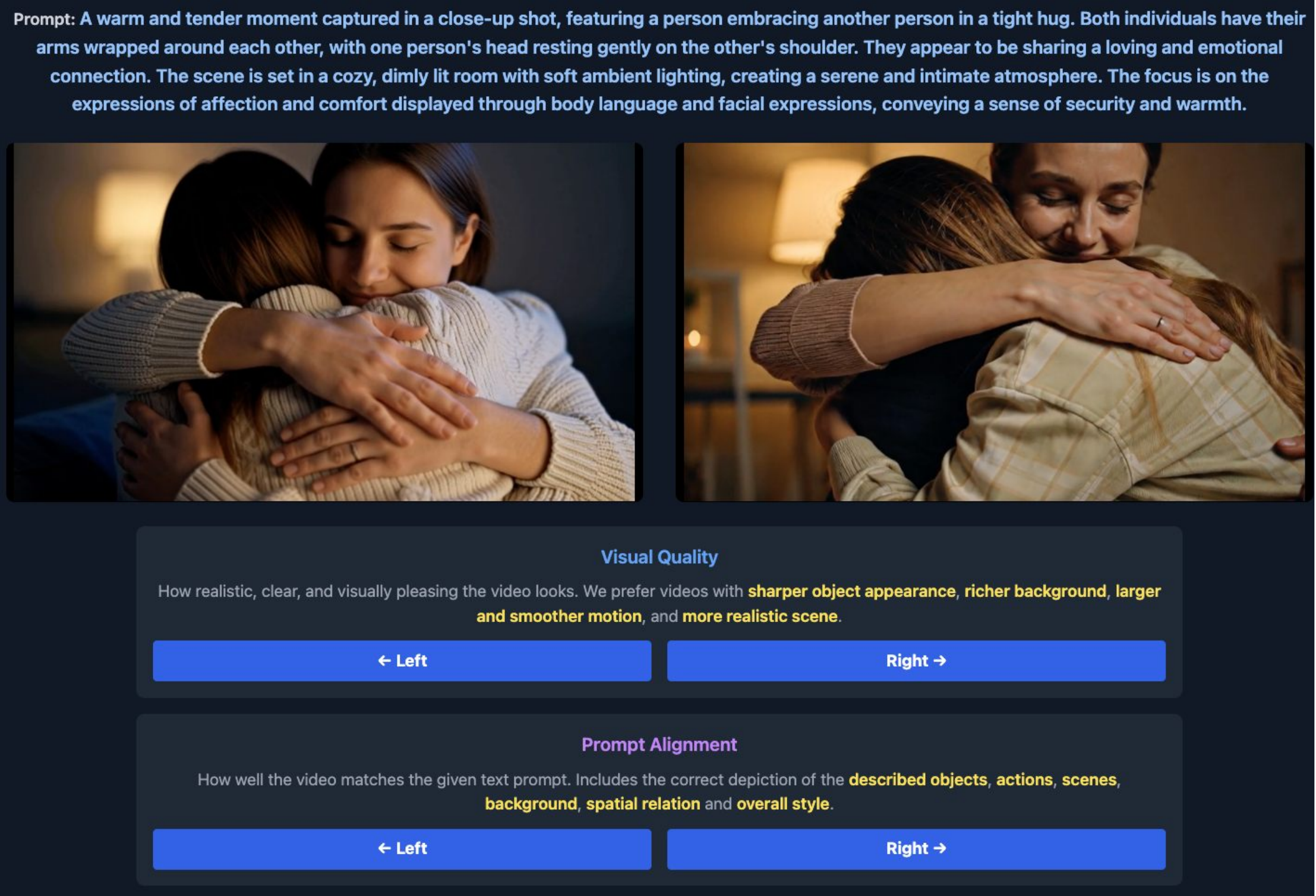}

    \vspace{-6pt}
    \caption{\textbf{User study interface.} Screenshot of our user preference study interface explained in~\Cref{sec:comparison}.}
    \label{fig:user_study}
    \vspace{-5pt}
\end{figure*}

\end{document}